\documentclass[10pt,journal,compsoc]{IEEEtran}
\ifCLASSOPTIONcompsoc
  \usepackage[nocompress]{cite}
\else
  \usepackage{cite}
\fi

\usepackage[english]{babel}

\ifCLASSINFOpdf
  \usepackage[pdftex]{graphicx}
  \graphicspath{{./graphics/}}
\else
  \usepackage[dvips]{graphicx}
  \graphicspath{{./graphics/}}
\fi

\usepackage{amssymb}
\usepackage{amsthm}
\usepackage{amsmath}
\usepackage{color}
\usepackage[usenames,dvipsnames,svgnames,table]{xcolor}
\usepackage{booktabs}       % professional-quality tables
\usepackage[ruled,section]{algorithm}
\usepackage[noend]{algorithmic}
\usepackage{array}
\ifCLASSOPTIONcompsoc
  \usepackage[caption=false,font=normalsize,labelfont=sf,textfont=sf]{subfig}
\else
  \usepackage[caption=false,font=footnotesize]{subfig}
\fi

\usepackage{epstopdf}
\usepackage{url}
\usepackage{wrapfig}
\usepackage{tikz}
\usepackage{pgfplots}
\usetikzlibrary{arrows,positioning,automata,bayesnet}
\allowdisplaybreaks
\allowbreak

\DeclareMathOperator*{\argmax}{argmax}
\newtheorem{theorem}{Theorem}
\newtheorem{definition}{Definition}
\newtheorem{lemma}{Lemma}

\pgfplotsset{
    compat=newest,
    every axis/.append style={
        legend image post style={xscale=0.5}
    },
}

\urldef{\myurl}\url{https://www.cc.gatech.edu/%7Eborg/ijcv%5Fpsslds/}

\begin{document}

\title{Time Series Change Point Detection with Gaussian Processes}

% Time Series Change Point Detection with Gaussian Processes

\author{Olga Isupova,
        Danil Kuzin,
        Fredrik Gustafsson,~\IEEEmembership{Fellow,~IEEE}
        and~Lyudmila~Mihaylova,~\IEEEmembership{Senior Member,~IEEE}%
%\IEEEcompsocitemizethanks{\IEEEcompsocthanksitem O.Isupova is with the Department of Engineering Science, University of Oxford, UK.\protect\\
%E-mail: olga.isupova@eng.ox.ac.uk
%\IEEEcompsocthanksitem D.Kuzin is with Physics Department, Lancaster University, UK.\protect\\
%E-mail: d.kuzin@lancaster.ac.uk
%\IEEEcompsocthanksitem D.Kuzin is with Physics Department, Lancaster University, UK.\protec\\
%E-mail: kuzind@lancaster.ac.uk
%\IEEEcompsocthanksitem L.Mihaylova is with the Department
%of Automatic Control and Systems Engineering, the University of Sheffield, UK.\protect\\
%E-mail: l.s.mihaylova@sheffield.ac.uk
%\IEEEcompsocthanksitem F.Gustafsson is with the Department of Electrical Engineering, Linkoping University, Sweden.\protect\\
%E-mail:  fredrik.gustafsson@liu.se}% <-this % stops an unwanted space
%\thanks{Manuscript received XXX; revised XXX.}
}

\IEEEtitleabstractindextext{%

\begin{abstract}
This paper proposes a novel general Gaussian process framework for change point detection in time series. A change is defined as an alteration of the hyperparameters of the Gaussian process prior. We formulate the problem in the context of statistical hypothesis testing and demonstrate its performance over both offline and online multiple change points detection. The approach provides theoretical guaranties of its performance. An important advantage of the developed framework is that it does not need a detailed specification of the time series function before and after a change. The hyperparameter values are learned from the data. 
Comprehensive evaluation of the proposed framework is provided both on synthetic and real data and its efficiency is demonstrated.
\end{abstract}

}

\maketitle

\IEEEdisplaynontitleabstractindextext

% \fi
\IEEEpeerreviewmaketitle

\IEEEraisesectionheading{\section{Introduction}\label{sec:introduction}}

% \IEEEPARstart{\mathbf{A}}{demo} file is ....
% \IEEEPARstart{\mathbf{A}}{}demo file is ....
\IEEEPARstart{C}{hange} point detection methods are aimed to detect abrupt variations in time series where time series reflect the system behaviour~\cite{aminikhanghahi2017survey}. 
Change point detection is applied in a wide range of fields: anomaly detection~\cite{Briassouli2011}; cyber-attack detection~\cite{Wang2004DoSChange}; finance and economics~\cite{Liu2010}; autonomous driving~\cite{Galceran2017}; motion segmentation~\cite{Zhou2013, Gong2014}; speech processing, analysis of biomedical signals, adaptive filtering and tracking~\cite{gustafsson2000adaptive, Lai1995, Wang2003PAMI, Han2004}.

%such that before a change the data follows one probability distribution and after the change it starts to follow another distribution. Change point detection is applied in a wide range of fields: anomaly detection~\cite{Briassouli2011};
%

Although introduced in 1954, the cumulative sum (CUSUM) algorithm~\cite{Page1954} remains a very popular method for change point detection in its original form or its extensions~\cite{Woodall1985, Lucas1982,Tartakovsky2006}. The CUSUM algorithm is an example of a stopping rule (or control chart) procedure developed for change point detection~\cite{Roberts1966}. However, the CUSUM algorithm can also be derived within a statistical hypothesis testing approach. Statistical hypothesis testing is an active area of change point detection where one can formulate a null hypothesis corresponding to a case without a change against the alternative that there is a change. A number of diverse algorithms within this approach rely on different assumptions about a change nature and data distribution~\cite{Tartakovsky2014, Lai2000, Ho2010}. Another direction in the change point detection area is the Bayesian approach where a posteriori probability of a change is computed~\cite{Shiryaev1961, Barry1993, Carlin1992, Zhitlukhin2013, aminikhanghahi2017survey}.  

The methods for change point detection can be classified into two categories: offline and online. Offline methods require a whole dataset of observations, e.g., change point locations can be found by minimising a global penalty function~\cite{Lavielle2005, Lebarbier2005, Bai1997}. In contrast, online methods process data sequentially and generate alarms about detected changes, e.g., by applying the full Bayesian inference~\cite{Adams2007}, or within a filtering framework~\cite{Azimi2002}, or by comparing dissimilarity between two subsets of data indicating a change time between these two subsets~\cite{Desobry2005}.  

Changes in time series can be considered as functional switches where a data stream follows one functional dependence before a change and another functional dependence expresses the data generative model after the change. From this perspective Gaussian processes represent a promising methodology for change point detection as they are often used as a prior on functions~\cite{Kim2006,Williams1998,Filippone2014,Bratieres2015}. Gaussian process is a generalisation of a Gaussian distribution into an infinite continuous domain, i.e., functions~\cite{rasmussen2006gaussian}.

In this paper changes are considered as functional breaks in time series. Functions governing observed data have a Gaussian process prior and a change in time series is defined as a change in hyperparameters of this Gaussian process. We employ a statistical hypothesis testing framework and propose statistical tests for change point detection.

Gaussian processes are explored for change point detection in literature. In~\cite{Chandola2011} a Gaussian process is used to perform a one-step ahead prediction and a control chart is used to detect a change point based on the difference between a prediction and an actual observation. Being an online method, it uses the whole historical data to make a prediction. However, inference in the Gaussian process framework faces scalability challenges  with the growth of the training datasets. The use of all historical data also adds a difficulty to the ability of the algorithm to adapt to different stationary regimes to detect multiple changes.

An expected value of observed data is modelled with a Gaussian process in~\cite{Grande2017} and changes in the mean are detected with the likelihood ratio test. Instead of  detecting a single point that deviates from a normal data flow, the method proposed in~\cite{herlands2018gp_subset} detects an anomalous subset of data assuming that data comes from a Gaussian process prior. Both~\cite{Grande2017, herlands2018gp_subset} detect changes only as a mean shift and do not consider changes in a covariance structure.

Change points as changes in hyperparameters of a Gaussian process prior are considered in~\cite{Saatci2010, Garnett2009, herlands2016scalable}. In~\cite{Garnett2009} change locations are incorporated into a covariance function and in~\cite{herlands2016scalable} data is represented as a mixture of functions from different Gaussian process priors where the weights of the mixture indicate a change. These methods work within the Bayesian approach and estimate the posterior probability of a change. Following the Bayesian approach unknown hyperparameters are  marginalised, although the obtained integrals are intractable. Therefore, these integrals need to be approximated: e.g., by a grid-based method or Hamiltonian Monte Carlo~\cite{Saatci2010} and by Bayesian Monte Carlo~\cite{Garnett2009}. The method in~\cite{herlands2016scalable} is also limited to offline detection. 
In contrast to the above mentioned methods, the proposed approach in the context of statistical hypothesis testing provides theoretical guaranties of its performance. The proposed methods rely on exact inference for likelihood and hyperparameter estimation.

The statistical hypothesis testing method for change point detection in Gaussian process data is proposed in~\cite{Keshavarz2015}. In that work a change is considered in a mean value of observed data. The method is offline and designed for detection of a single change point. In contrast to this method, we consider changes in any of the hyperparameters of the Gaussian process, including the ones of mean and covariance functions. The proposed methods are suitable for both offline and online data processing and multiple change points detection.  

This work has the following main contributions: \emph{i}) The change point detection problem is formulated within the statistical hypothesis testing framework, where a change is defined as a change in hyperparameters of a Gaussian process prior.  \emph{ii}) A general framework for change point detection is proposed. \emph{iii}) Statistical tests for change point detection are defined and developed. Theoretical properties of the introduced statistical tests are derived. \emph{iv}) A change point detection approach for online data processing able to detect multiple change points is designed. \emph{v}) The proposed approach is thoroughly evaluated in terms of the typical change point detection measures on synthetic and real data. 

The part of this work has appeared in a PhD thesis chapter by the first author~\cite{Isupova2018}. In constrast to~\cite{Isupova2018}, in this work we additionally propose a novel method for online change point detection with estimated hyperparameters, evaluate sensitivity of a parameter used for hyperparameter optimisation, analyse computational complexity of the proposed methods  and perform more detailed experiments.  

The rest of the paper is organised as follows. Section~\ref{sec:formulation} formulates a change point detection problem in 	Gaussian process time series. Section~\ref{sec:lrt} describes the proposed likelihood ratio tests. Online statistical tests are introduced in Section~\ref{sec:online_tests}. Section~\ref{sec:complexity} discusses computational complexity of the proposed tests. The proofs of the main theoretical results are presented in Section~\ref{sec:proofs}. The proposed approaches are evaluated on simulated data in Section~\ref{sec:synthetic} and on real data in Section~\ref{sec:real_data}. Section~\ref{sec:conclusion} concludes the paper.


\section{Problem formulation}
\label{sec:formulation}

Change point detection aims to detect abrupt changes in time series data. An abrupt change is understood as a change in a latent probability distribution of observed data. 

\subsection{Data model}
We consider the following Gaussian process (GP) time series model:
\begin{equation}
\label{eq:data_model}
y_t = f(t) + \varepsilon_t,
\end{equation}
where $y_t\in\mathbb{R}$ is an observation at a time moment~$t$; $f(\cdot) \sim \mathcal{GP}(m(\cdot), k(\cdot, \cdot))$ is a function of time with a GP prior, where $\mathcal{GP}$ denotes a GP, characterised by a mean function~$m(\cdot)$ and a covariance function $k(\cdot, \cdot)$; $\varepsilon_t \sim \mathcal{N}(\varepsilon_t | 0, \sigma^2)$ is a Gaussian noise with a zero mean and a variance~$\sigma^2$. 
The mean and covariance functions of the GP are parameterised with vectors $\boldsymbol\vartheta^{m}$ and $\boldsymbol\vartheta^{k}$, respectively. Then $\boldsymbol\vartheta = \{\boldsymbol\vartheta^{m}, \boldsymbol\vartheta^{k}, \sigma^2\}$ denotes a vector of hyperparameters of a given GP.
 
%Although in this paper we consider only the GP time series model, the proposed methods are directly applied to the GP autoregressive model:
The proposed approach is valid for the GP autoregressive model:
\begin{equation*}
y_t = f(y_{t - r}, \dotsc, y_{t - 1}) + \varepsilon_t,
\end{equation*}
where $r$ is an order of the model. However, for simplicity, we will consider only the GP time series model~(\ref{eq:data_model}) in this paper.

For a given set of $N$ input points of time indices $\mathbf{t}_{1:N} = [t_i]_{i = 1}^N$ we can compute the distributions of function values and observations~\cite{rasmussen2006gaussian}:
\begin{equation}
\mathbf{f}_{1:N} | \mathbf{t}_{1:N} \sim \mathcal{N}(\mathbf{f}_{1:N} | \boldsymbol{\mu}, \mathbf{K}),
\end{equation}
where $\mathbf{f}_{1:N} = f(\mathbf{t}_{1:N}) = [f(t_i)]_{i = 1}^{N}$ are values of function $f$ for the given input time indices; $\boldsymbol\mu = [\mu_i]_{i = 1}^{N} = [m(t_i)]_{i = 1}^N$ are realisations of the GP mean function in the input points;
$\mathbf{K} = \left[\mathbf{K}_{i, j}\right]_{i, j = 1}^N = \left[k(t_i, t_j)\right]_{i, j = 1}^N$ are realisations of the GP covariance function in the input points; and
\begin{equation}
\mathbf{y}_{1:N} | \mathbf{t}_{1:N} \sim \mathcal{N}(\mathbf{y}_{1:N} | \boldsymbol{\mu}, \mathbf{K} + \sigma^2 \mathbf{I}),
\end{equation}
where $\mathbf{y}_{1:N} = [y_i]_{i = 1}^N$ are observations at the given time indices and $\mathbf{I}$ is the identity matrix\footnote{All vector variables are assumed to be column-vectors. Vector variables are denoted as bold lower case letters.}.

The marginal log likelihood function of observed data is
\begin{equation}
\label{eq:log_lik}
\begin{split}
\log p(&\mathbf{y}_{1:N} | \mathbf{t}_{1:N}, \boldsymbol\vartheta) \\
= & -\dfrac{1}{2} \left(\mathbf{y}_{1:N} - \boldsymbol\mu\right)^{T} \left(\mathbf{K} + \sigma^2 \mathbf{I}\right)^{-1} (\mathbf{y}_{1:N} - \boldsymbol\mu) \\
& - \dfrac{1}{2} \log |\mathbf{K} + \sigma^2 \mathbf{I} | - \dfrac{N}{2} \log 2 \pi,
\end{split}
\end{equation}
where $|\cdot|$ is a determinant of a matrix.

\subsection{Change point detection problem formulation}
An abrupt change in GP time series data can be defined as a change in hyperparameters of the GP. This means that the function $f$ from the data model~(\ref{eq:data_model}) has a GP prior governed by different hyperparameters before and after the change. The vectors of hyperparameters before and after the change are denoted as $\boldsymbol\vartheta_0$ and $\boldsymbol\vartheta_1$, respectively. The final model for time series data with an abrupt change is then
\begin{subequations}
\begin{align}
\label{eq:change_data_model_1}
y_t &= f_0(t) + \varepsilon_{t}^0, \qquad \forall t < t^{*};\\
\label{eq:change_data_model_2}
y_t &= f_1(t) + \varepsilon_{t}^1, \qquad \forall t \geq t^{*},
\end{align}
\label{eq:change_data_model}
\end{subequations}
where $t^{*}$ is a time instant when the change occurs and
\begin{subequations}
\begin{align}
f_0(\cdot) &\sim \mathcal{GP}(m_{0}(\cdot), k_{0}(\cdot, \cdot)); &\varepsilon_t^0 &\sim \mathcal{N}(0, \sigma_{0}^2);\\
f_1(\cdot) &\sim \mathcal{GP}(m_{1}(\cdot), k_{1}(\cdot, \cdot)); &\varepsilon_t^1 &\sim \mathcal{N}(0, \sigma_1^2),
\end{align}
\end{subequations}
where $m_{a}$ and $k_{a}$ are the mean and covariance functions of the GP governed by the hyperparameter vector $\boldsymbol\vartheta_a$, $\sigma_{a}^2$ is the variance of the additive noise defined by the hyperparameter vector $\boldsymbol\vartheta_a$, for $a \in \{0, 1\}$.

The change point detection problem can be formulated as a statistical hypothesis testing task. Let the null hypothesis~$\mathcal{H}_0$ state that a GP prior of a function~$f$ has a hyperparameter vector~$\boldsymbol\vartheta$ equal to the initial value $\boldsymbol\vartheta_0$ during the whole observation period. Meanwhile let the alternative~$\mathcal{H}_1$ claim that there exists a change time index~$t^{*}$ such that the hyperparameter vector~$\boldsymbol\vartheta$ of the GP has the initial value~$\boldsymbol\vartheta_0$ before the change point and is equal to a new vector~$\boldsymbol\vartheta_1$ after the change time~$t^{*}$:   
\begin{subequations}
\begin{align}
\label{eq:H_0}
\mathcal{H}_0: \qquad&\quad\boldsymbol\vartheta = \boldsymbol\vartheta_0, \quad \forall t;\\
\label{eq:H_1_1}
\mathcal{H}_1: \exists t^{*}: &\begin{cases}\boldsymbol\vartheta = \boldsymbol\vartheta_0, \quad t < t^{*},\\
\boldsymbol\vartheta = \boldsymbol\vartheta_1, \quad t \geq t^{*}.
\end{cases}
\end{align}
\label{eq:Hs}
\end{subequations}

The next sections provide a description of the proposed statistical methods	for the testing problem introduced in~(\ref{eq:Hs}).

\section{The Gaussian process change point detection based on likelihood ratio tests}
\label{sec:lrt}
This section introduces offline change point detection methods that given the whole time series find a possible change point in it. The methods use likelihood ratio based tests since the likelihood ratio test (LRT) is the optimal test according to the Neyman-Pearson lemma. We start with the case when both hyperparameters $\boldsymbol\vartheta_0$ and $\boldsymbol\vartheta_1$ are assumed to be known (Section~\ref{subsec:lrt}) and then relax this assumption (Section~\ref{subsec:gen_lrt}).

%This section introduces two likelihood ratio based tests for the change point detection problem~(\ref{eq:Hs}). The first proposed test is the pure likelihood ratio test (LRT) for the considered problem. In this set-up both hyperparameter values $\boldsymbol\vartheta_0$ and $\boldsymbol\vartheta_1$ are assumed to be known. For some of the real world applications this assumption is unlikely to hold; therefore, the second proposed test is the generalised LRT (gLRT), where the hyperparameter vectors $\boldsymbol\vartheta_0$ and $\boldsymbol\vartheta_1$ are estimated from the data.  

\subsection{Likelihood ratio test}
\label{subsec:lrt}
Let $\mathbf{y}_{1:N}$ be observed output variables, $\boldsymbol\vartheta_0$ and $\boldsymbol\vartheta_1$ are known hyperparameter vectors of the GP before and after a change, respectively. We propose a likelihood ratio test for this case. Define a LRT statistic for change point detection at a time moment $t^{*}$ as the logarithm of ratio of the likelihood under the assumption that the change occurs at the time moment $t^{*}$ and the likelihood of the data without the change. Formally, the LRT statistic $T_{\text{LRT}}(t^{*}|\mathbf{y}_{1:N})$ is defined as
\begin{align}
T_{\text{LRT}}&(t^{*} | \mathbf{y}_{1:N}) \nonumber\\
\stackrel{\text{def}}{=} &2\log\left(\dfrac{p(\mathbf{y}_{1:t^{*}-1}|\mathbf{t}_{1:t^{*}-1}, \boldsymbol\vartheta_0) p(\mathbf{y}_{t^{*}:N}|\mathbf{t}_{t^{*}:N}, \boldsymbol\vartheta_1)}{p(\mathbf{y}_{1:N}|\mathbf{t}_{1:N}, \boldsymbol\vartheta_0) }\right) \nonumber\\
= & 2\log p(\mathbf{y}_{1:t^{*}-1}|\mathbf{t}_{1:t^{*}-1}, \boldsymbol\vartheta_0) + 2\log p(\mathbf{y}_{t^{*}:N}|\mathbf{t}_{t^{*}:N}, \boldsymbol\vartheta_1) \nonumber\\
\label{eq:lrt_stat}
& - 2\log p(\mathbf{y}_{1:N}|\mathbf{t}_{1:N}, \boldsymbol\vartheta_0),
\end{align} 
where $\log p(\mathbf{y}_{1:t^{*}-1}|\mathbf{t}_{1:t^{*}-1}, \boldsymbol\vartheta_0)$ is the log likelihood of the data starting from the first time moment till the time moment $t^{*}-1$ computed based on the GP hyperparameter vector $\boldsymbol\vartheta_0$; $\log p(\mathbf{y}_{t^{*}:N}|\mathbf{t}_{t^{*}:N}, \boldsymbol\vartheta_1)$ is the likelihood of the data starting at the time moment $t^{*}$ till the end of the observation period given the hyperparameter vector $\boldsymbol\vartheta_1$; $\log p(\mathbf{y}_{1:N}|\mathbf{t}_{1:N}, \boldsymbol\vartheta_0)$ is the likelihood of the whole set of observations given the hyperparameter vector $\boldsymbol\vartheta_0$. The log likelihood terms can be obtained by (\ref{eq:log_lik}).

\begin{definition}
\label{def:lrt_test}
\textbf{The likelihood ratio test for change point detection} is defined as:
\begin{equation}
  \max_{t^{*}} T_{\text{LRT}}(t^{*}|\mathbf{y}_{1:N}) > c_{\text{LRT,\,THR}},
\end{equation}
where $T_{\text{LRT}}$ is given by (\ref{eq:lrt_stat}) and $c_{\text{LRT,\,THR}}$ is a threshold.
If the threshold $c_{\text{LRT,\,THR}}$ is exceeded, the time moment~$t^{*}$ that maximises~$T_{\text{LRT}}$ is chosen as
the estimated change time.
\end{definition}

For any given $t^{*}$, the LRT is the optimal test according to the Neyman-Pearson lemma~\cite{gustafsson2000adaptive}. 

The threshold $c_{\text{LRT,\,THR}}$ value should be positive, as the statistic value $T_{\text{LRT}}(N|\mathbf{y}_{1:N})$ of the last time moment is equal to zero and this case by convention corresponds to the ``no change'' decision by the test. 

\subsection{Generalised likelihood ratio test}
\label{subsec:gen_lrt}
Here we relax the assumption that the hyperparameters of the GP are known with the proposed generalised LRT (GLRT). 
The GLRT is used as the LRT (Definition~\ref{def:lrt_test}), but values for the vectors $\boldsymbol\vartheta_0$ and $\boldsymbol\vartheta_1$ are estimated from the data. The GLRT statistic is given by:
\begin{align}
T_{\text{GLRT}}&(t^{*}|\mathbf{y}_{1:N}) \nonumber\\
 \stackrel{\text{def}}{=} & 2\log\left(\frac{p(\mathbf{y}_{1:t^{*}}|\mathbf{t}_{1:t^{*}}, \hat{\boldsymbol\vartheta}_0) p(\mathbf{y}_{t^{*}+1:N}| \mathbf{t}_{t^{*}+1:N}, \hat{\boldsymbol\vartheta}_1)}{p(\mathbf{y}_{1:N}|\mathbf{t}_{1:N}, \hat{\boldsymbol\vartheta}_0)}\right),
\end{align}
where $\hat{\boldsymbol\vartheta}_0$ and $\hat{\boldsymbol\vartheta}_1$ are estimates of the hyperparameter vectors of the GP.

The estimates of the hyperparameters can be obtained by maximising the marginal likelihood~(\ref{eq:log_lik})~\cite{rasmussen2006gaussian}:
\begin{align}
\hat{\boldsymbol\vartheta}_a = & \argmax\limits_{\boldsymbol\vartheta_a} \log p(\mathbf{y} | \mathbf{t}, \boldsymbol\vartheta_a), \quad a \in \{0, 1\}.
\end{align}

Note that we cannot formally separate estimation of the change point and estimation of the parameters $\boldsymbol\vartheta_0$ and $\boldsymbol\vartheta_1$ and normally a generalised LRT solves the double maximisation problem: over location of a change point and parameter values. In our experiments with this test we use $M$ first and $M$ last data points of the whole sequence $\mathbf{y}_{1:N}$, assuming that the change occurs somewhere inside the sequence: $t^{*} \in \{M+1, \ldots, N - M\}$.

\subsection{Discussions} 
The likelihood ratio tests represent the core elements in change point detection and realisations of these tests for different problems are widely used in literature~\cite{gustafsson2000adaptive}. These tests are simple to interpret; they are easily implemented. The likelihood ratio test with known hyperparameters is proven to be optimal. The generalised likelihood ratio test uses the estimates of the hyperparameters and it might be expected to be near optimal if the estimates of the hyperparameters are close enough to the true unknown values.

However, in the real-world applications a practitioner might face some issues using the likelihood ratio tests. Firstly, tests are designed for offline data processing. It can be inappropriate for settings when a decision should be made in an online manner and some actions are expected in the case of an alarm about a change.

Secondly, the tests expect no more than one change point within the whole processed dataset. Even if online processing is not essential, this limitation to have no more than one change point implies further constraints on applicability of the tests. 

In order to overcome these issues another family of statistical tests is proposed in the next section. These tests work within a sliding window allowing both online data processing and multiple change points detection. 

\section{The Gaussian process online change point detection based on likelihood estimation}
\label{sec:online_tests}
This section introduces online change point detection methods that process sequentially upcoming data. Let $\mathbf{y}_{1:t}$ be observed data up to the current moment $t$ that follows the model~(\ref{eq:change_data_model}). At each time moment $t$ we are interested if there is a change in a hyperparameter vector $\boldsymbol\vartheta$ from its initial value $\boldsymbol\vartheta_0$. Note that the new value $\boldsymbol\vartheta_1$ is neither known nor estimated from the data to make a decision about the change. After the change has been detected, it may be relevant to estimate the new set of hyperparameters, and then restart the test to look for the next change. As in Section~\ref{sec:lrt} we start with the case when $\boldsymbol\vartheta_0$ is assumed to be known (sections~\ref{subsec:online_test_known}~--~\ref{subsec:online_test_theory}) and then relax this assumption (Section~\ref{subsec:online_test_est}).

\subsection{Test formulation}
\label{subsec:online_test_known}
In this section we propose a test that process a data within a sliding window $\mathbf{y}_{t-L+1:t}$ of length $L$. For the test we compute the following statistic
\begin{align}
T&_{\text{online}}(\mathbf{y}_{t - L + 1:t}) \stackrel{\text{def}}{=} -2 \log p(\mathbf{y}_{t-L+1:t} | \mathbf{t}_{t-L+1:t}, \boldsymbol\vartheta_0) \nonumber\\
& - \log | \mathbf{K}_{0} + \sigma_{0}^2 \mathbf{I} | - L \log 2\pi \nonumber\\
\label{eq:online_stat}
= & \left(\mathbf{y}_{t-L+1:t} - \boldsymbol\mu_0\right)^{T} \left(\mathbf{K}_0 + \sigma_0^2 \mathbf{I}\right)^{-1} \left(\mathbf{y}_{t-L+1:t} - \boldsymbol\mu_0\right).
\end{align}

%Note that $\log p(\mathbf{y}_{t-L+1:t} | \mathbf{t}_{1:t}, \boldsymbol\vartheta) \equiv \log p(\mathbf{y}_{t-L+1:t} | \mathbf{t}_{t-L+1:t}, \boldsymbol\vartheta)$, i.e., the likelihood of the data within the sliding window depends only on the time indices within this sliding window.

\begin{theorem}
\label{th:online_distr}
Consider the problem (\ref{eq:Hs}), where $\mathbf{y}_{1:t}$ is the observed data up to the time moment $t$ and $\boldsymbol\vartheta_0$ is a known hyperparameter vector of the GP before a change. Let $T_{\text{online}}(\mathbf{y}_{t - L + 1:t})$ be the likelihood-based statistic defined as (\ref{eq:online_stat}). Then given that the null hypothesis $\mathcal{H}_0$ is true:
\begin{equation}
T_{\text{online}}(\mathbf{y}_{t - L + 1:t}) \sim \chi^{2}_L,
\end{equation}
where $\chi^2_L$ is a chi-squared distribution with $L$ degrees of freedom. 
\end{theorem}
Proof of Theorem \ref{th:online_distr} is presented in Section \ref{subsec:stat_distr_th_proof}. 

\begin{figure}[!t]
\centering
\includegraphics[width = 0.35\textwidth]{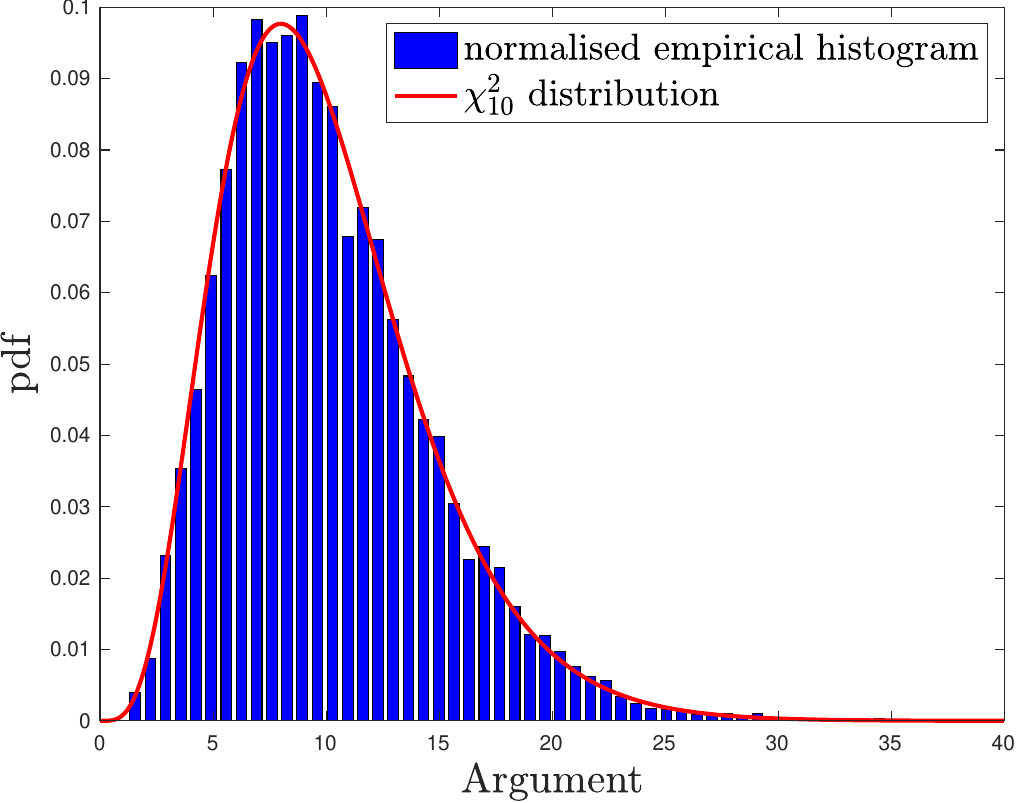}
\caption{Empirical normalised histogram of the test likelihood-based statistic with the corresponding analytical $\chi^2$ distribution.}
\label{fig:stat_emp}
\end{figure}

Figure~\ref{fig:stat_emp} shows a visual comparison of an empirical normalised histogram of the statistic $T_{\text{online}}$ and the analytical $\chi^{2}_L$ distribution. The normalised histogram is built based on $10\,000$ Monte Carlo simulations with $\mathbf{y} \sim \mathcal{N}\left(\mathbf{y} | \boldsymbol\mu_{0} = \mathbf{0}, \mathbf{K}_{0}\right)$, where $\mathbf{K}_{0}$ is a squared exponential covariance matrix: 
\begin{equation}
\mathbf{K}_{0}(i, j) = \sigma_\mathbf{K}^2 \exp\left(-\dfrac{(t_i - t_j)^2}{2 \ell^2}\right)
\end{equation}
with the signal variance $\sigma_\mathbf{K}^2 = 0.5$ and the length scale $\ell = 1$. The length of the vector $\mathbf{y}$ is $10$. Noise-free observations are considered, therefore $\sigma_{0}^2 = 0$. The Kolmogorov-Smirnov test does not reject a hypothesis that the sample of $T_{\text{online}}$ is obtained from the corresponding $\chi^{2}_{10}$ distribution with a significance level $0.05$ and $p$-value of~$0.495$. 

Once the statistic distribution under the null hypothesis is determined the test procedure can be defined.
\begin{definition}
\label{def:online_test}
\textbf{The online likelihood-based test for change point detection} is defined as:
Let $T_{\text{online}}(\mathbf{y}_{t-L+1:t})$ be a statistic given in~(\ref{eq:online_stat}), which is $\chi^2_L$-distributed under the null hypothesis $\mathcal{H}_0$. Let $e_1$ and $e_2$ be quantiles of the $\chi^2_L$ distribution such that $\mathcal{F}_{\chi^2_L}(e_1) = \frac{\alpha_{\text{stat}}}{2}$ and $\mathcal{F}_{\chi^2_L}(e_2) = 1 - \frac{\alpha_{\text{stat}}}{2}$, where $\mathcal{F}_{\chi^2_L}(\cdot)$ is the cumulative density function (cdf) of the $\chi^2_L$ distribution and $\alpha_{\text{stat}}$ is a given significance level. 
Reject $\mathcal{H}_0$, if $T_{\text{online}}(\mathbf{y}_{t-L+1:t}) < e_1$  or $T_{\text{online}}(\mathbf{y}_{t-L+1:t}) > e_2$.
\end{definition}

\subsection{Theoretical evaluation of the test}
\label{subsec:online_test_theory}
Effectiveness of a statistical test is usually evaluated by probabilities of type-I and type-II errors. A type-I error is a rejection of the null hypothesis when it is true (i.e., a false alarm), while a type-II is a failure to reject the null hypothesis while the alternative is true (i.e., a missed detection). The probability of the type-I error for the proposed test is fixed and equal to $\alpha_{\text{stat}}$. We do not evaluate the probability $\beta_{\text{stat}}$ of the type-II error directly, we rather evaluate the power of the test, which is a complement to $\beta_{\text{stat}}$.

The power $B(\boldsymbol\vartheta)$ of the test is defined as~\cite{Basseville1993}:
\begin{equation}
\label{eq:power_def}
B(\boldsymbol\vartheta) \stackrel{\text{def}}{=} \mathbb{P}(\text{reject } \mathcal{H}_0 | \mathcal{H}_1\text{ is true}).
\end{equation}

The following distribution is useful in the context of the proposed test and its power:

\begin{definition}
\label{def:gen_chi_sq}
Let $\zeta_i$ be independent random variables having non-central chi-squared distributions and $a_i$ be deterministic coefficients, $i \in \{1, \dotsc, I\}$. Then a random variable $\tilde{\zeta}$:
\begin{equation}
\tilde{\zeta} = \sum\limits_{i = 1}^I a_i \zeta_i
\end{equation}
has a \textbf{generalised chi-squared distribution}.
\end{definition}
A generalised chi-squared distribution has no closed-form expression but can be efficiently computed numerically~\cite{davies1980algorithm}.

\begin{theorem}
\label{th:power}
Consider the problem (\ref{eq:Hs}) and the online likelihood-based change point detection test defined in Definition~\ref{def:online_test}. Then the power of this test is:
\begin{equation}
B(\boldsymbol\vartheta) = 1 + \mathcal{F}_q(e_1) - \mathcal{F}_q(e_2), 
\end{equation}
where $\mathcal{F}_q(\cdot)$ is the cdf of the random variable $q$. The variable $q$ follows the generalised chi-squared distribution:
\begin{equation}
q = \sum\limits_{i = 1}^L \lambda_i v_i,
\end{equation}
where $\lambda_i$ are eigenvalues of the matrix $\mathbf{A}$:
\begin{equation}
\mathbf{A} = \left(\mathbf{K}_{1} + \sigma^2_{1} \mathbf{I} \right)^{\frac{1}{2}}
\left(\mathbf{K}_{0} + \sigma^2_{0} \mathbf{I} \right)^{-1} \left(\mathbf{K}_{1} + \sigma^2_{1} \mathbf{I} \right)^{\frac{1}{2}},
\end{equation} 
and $v_i$ are random variables: $v_i \sim \chi^{'2}_1(s_i^2)$ while $s_i$ are components of the vector $\mathbf{s} = \mathbf{P}^{T} \left(\mathbf{K}_{1} + \sigma^2_{1} \mathbf{I} \right)^{-\frac{1}{2}} \left(\boldsymbol\mu_{1} - \boldsymbol\mu_{0}\right)$, where $\mathbf{P}$ is a matrix, which columns are eigenvectors of the matrix $\mathbf{A}$.
\end{theorem}
Notation $\chi^{'2}_k(\lambda)$ is for a non-central chi-squared distribution with $k$ degrees of freedom and the noncentrality parameter $\lambda$.

Proof of the Theorem \ref{th:power} is presented in Section \ref{subsec:power_th_proof}.

Theorem~\ref{th:power} allows to calculate the theoretical success rate of test application for any given data distribution hyperparameters $\boldsymbol\vartheta_1$.

The sliding window nature of the proposed test (Definition~\ref{def:online_test}) makes it applicable for online data processing. In practice the true hyperparameter vector~$\boldsymbol\vartheta_0$, which can be unknown, may be replaced by its estimate. Note, that the data that is used for hyperparameter vector estimation is not limited to the sliding window employed in the test statistic computation. More historical data can be used, therefore an accurate estimate might be expected. %Besides, if the estimates of the hyperparameters are updated based on recent data the test would be able to detect multiple changes between different data distributions.  

\subsection{Test with estimated hyperparameters}
\label{subsec:online_test_est}
Consider the online likelihood-based change point detection test (Definition~\ref{def:online_test}) when the initial hyperparameter vector~$\boldsymbol\vartheta_0$ is unknown. The value of $\boldsymbol\vartheta_0$ can be estimated using previously observed data $\mathbf{y}_{t':t''}$, where $t' \leq t'' < t$, for example, by maximising the marginal likelihood:
\begin{equation}
\hat{\boldsymbol\vartheta}_0 = \argmax\limits_{\boldsymbol\vartheta_0} \log p(\mathbf{y}_{t':t''} | \mathbf{t}_{t':t''}, \boldsymbol\vartheta_0).
\end{equation} 
The test statistic $T_{\text{online est}}$ for the online likelihood-based change point detection test with estimated hyperparameters is defined as
\begin{equation}
\begin{split}
T_{\text{online est}}&(\mathbf{y}_{t - L + 1:t})  \stackrel{\text{def}}{=} -2 \log p(\mathbf{y}_{t-L+1:t} | \mathbf{t}_{t-L+1:t}, \hat{\boldsymbol\vartheta}_0)\\
& - \log | \hat{\mathbf{K}}_{0} + \hat{\sigma}_{0}^2 \mathbf{I} | - L \log 2\pi,
\end{split}
\end{equation}
where $\hat{\mathbf{K}}_{0}$ and $\hat{\sigma}_{0}^2$ are based on $\hat{\boldsymbol\vartheta}_0$.

In experiments we consider two options for choosing $t'$ and $t''$.
\subsubsection{Sliding window for optimisation}
\label{subsubsec:sliding_window_opt}
Here $t'$ and $t''$ are set to form another sliding window of length $\tilde{L}$: $t' = t - \tilde{L}$ and $t'' = t-1$.

In such settings the estimate $\hat{\boldsymbol\vartheta}_0$ of the hyperparameter vector is always updated based on the most recent data and the test can be applied in an online manner. Moreover, it allows to adapt the test to different data distributions and the test is able to detect multiple changes between the periods with stationary data distributions, i.e., when the data model is:
\begin{subequations}
\begin{align}
y_t &= f_0(t) + \varepsilon_t^0, &\forall &t < t^{*}_1, &\boldsymbol\vartheta &= \boldsymbol\vartheta_0;\\
y_t &= f_1(t) + \varepsilon_t^1, &\forall t^{*}_1 \leq &t < t^{*}_2, &\boldsymbol\vartheta &= \boldsymbol\vartheta_1;\\
y_t &= f_2(t) + \varepsilon_t^2, &\forall t^{*}_2 \leq &t < t^{*}_3, &\boldsymbol\vartheta &= \boldsymbol\vartheta_2;\\
&\cdots \nonumber 
\end{align}
\label{eq:several_change_points}
\end{subequations}
and the goal is to detect all change points $t^{*}_1$, $t^{*}_2$, $t^{*}_3$, $\dotsc$ Due to the sliding window approach the test after the change point detection adapts to new data, enables to find a new change point breaking the new data stationary regime.

However, optimising hyperparameters at each time moment can be time-consuming. Therefore, we consider the second option of choosing the previosly observed data for hyperparameter optimisation.  

\subsubsection{Adaptive optimisation}
\label{subsubsec:adaptive_opt}
In this case $t'$ and $t''$ are set such that the hyperparameter vector $\boldsymbol\vartheta_0$ is estimated on every first $\tilde{L}$ data points after a change is detected (or at the very first observed data $\mathbf{y}_{1:\tilde{L}}$). Formally, we consider the set of $t'$ and $t''$ as
\begin{subequations}
\begin{align}
t' = 1 &\,\&\, t'' = \tilde{L};\\
t' = \hat{t}^{*}_1 &\,\&\, t'' = \hat{t}^{*}_1 + \tilde{L};\\
t' = \hat{t}^{*}_2 &\,\&\, t'' = \hat{t}^{*}_2 + \tilde{L};\\
&\cdots \nonumber
\end{align}
\end{subequations}
where $\hat{t}^{*}_1$, $\hat{t}^{*}_2$, $\dotsc$ are detected change points.

With these settings the test still can be applied in an online manner and for detecting several change points (\ref{eq:several_change_points}). In contrast to the sliding window approach with $t' = t - \tilde{L}$ and $t'' = t-1$, it requires to optimise hyperparameters only after change points. This approach needs to collect the first $\tilde{L}$ data points after a change before it can detect a new change point, i.e., change points $t^{*}_1$, $t^{*}_2$, $t^{*}_3$, $\dotsc$ should be at least $\tilde{L}$ apart to be detected. However, this limitation is reasonable if we consider $\tilde{L}$ as a minimum number of data points to determine a sound estimate of $\boldsymbol\vartheta_0$.

\subsection{Discussions}
The tests proposed in this section provide a tool for online change point detection. As a GP represents a prior for a wide variety of functions the proposed approach is general and can be applied in very different scenarios. For example, when data is expected to follow a smooth function a squared exponential covariance function can be used. If data is affected by seasonal variations, periodic covariance functions can come into play. 

To use the proposed tests the particular form of a GP specified by a choice of a hyperparameter vector is required only for data before a change. No assumptions about the distribution of data after a change are necessary. If data starts to follow a new rule, e.g., a different covariance function is used in a GP, the method will detect a change. Although for multiple change points detection in piecewise stationary data a structure of GPs for all stationary periods should be specified, but the exact values of hyperparameters would be estimated from the data by the algorithm.

The proposed online likelihood-based tests can be seen as the Bayesian information criterion (BIC) of the data model before a change. The term determing  of the BIC is constant in our settings. Note that in contrast to Bayesian model selection the proposed tests require knowledge about the form of the data model only before the change, while Bayesian model selection is based on comparison of models. For example, if we use the BIC for model selection we would need to compare the BIC values for the null hypothesis and alternative, i.e., we would need to determine a data model after a change. Moreover, proper Bayesian model selection requires marginalisation of hyperparameters which is intractable~\cite{Saatci2010, Garnett2009}.

Lengths of the data windows $L$ and $\tilde{L}$ are parameters of the tests. The smaller $L$ the faster a change can be detected since in a wider sliding window more points after a change are required to influence a joint probability of the points inside the window. However, too short window can produce noisy estimates. The length $\tilde{L}$ of the window for parameter estimation specify a trade-off between accuracy of estimates and speed of the algorithm as conventional computational complexity of parameter estimation scales cubically for GPs. In our experiments we find that relatively small values for both lengths provide satisfactory results: $L \in [3, \ldots, 10]$ and $\tilde{L} \in [15, \ldots, 40]$. Detailed analysis of accuracy of hyperparameter estimates with respect to $\tilde{L}$ is given in Appendix A. 

\section{Computational complexity of the proposed tests}
\label{sec:complexity}
All tests proposed in Sections~\ref{sec:lrt} and~\ref{sec:online_tests} rely on computation of the marginal log likelihood of an observation sequence~(\ref{eq:log_lik}), which is cubic with respect to the length of the sequence. The tests that use estimates of hyperparameters (Sections~\ref{subsec:gen_lrt} and~\ref{subsec:online_test_est}) also need to compute these estimates which have cubic complexity with respect to the length of a data sequence used for estimation computation. Therefore, the computational complexity of the proposed tests is:
\begin{enumerate}
\item LRT: $\mathcal{O}(N^3)$;
\item GLRT: $\mathcal{O}(N^3)$;
\item online likelihood-based change point detection test: $\mathcal{O}(N L^3)$ since the test is applied at each time moment;
\item online likelihood-based change point detection test with estimated hyperparameters: $\mathcal{O}(N \max\{L, \tilde{L}\}^3)$.
\end{enumerate}
The cubic complexity comes from inversion of a covariance matrix that is the most demanding operator. Note that the cubic complexity of GPs can be relaxed by using only a subset of observations (so-called inducing points)~\cite{Kuzin2018GPEnKF} or by using the Kalman filter reformulation of GP regression~\cite{sarkka2012infinite} that has a linear complexity. Since in our experiments we use relatively small values for both $L$ and $\tilde{L}$, different methods for reducing the cubic complexity of GPs are left out of the scope of the paper.

\section{Proofs of the theorems}
\label{sec:proofs}
This section presents the detailed proofs of the theorems introduced in Section~\ref{sec:online_tests}.

We start from auxiliary results regarding distributions of quadratic forms of random vectors. These results are then used in the proofs of the main theorems.

Consider a distribution of a quadratic form $\mathbf{y}^{T} \mathbf{K}^{-1} \mathbf{y}$, where $\mathbf{y} \in \mathbb{R}^n$ is a random vector distributed as multivariate Gaussian and $\mathbf{K} \in \mathbb{R}^{n \times n}$ is a deterministic matrix. Two cases are particularly interesting in the context of this paper: the case when $\mathbf{K}$ is a covariance matrix of the random vector $\mathbf{y}$ and the case when $\mathbf{K}$ is an arbitrary symmetric matrix.

\subsection{Quadratic form of the ``own'' covariance matrix}
Let $\mathbf{y} \sim \mathcal{N}(\mathbf{y} | \boldsymbol\mu, \mathbf{K})$, i.e., we are interested how the quadratic form of the covariance matrix of the given random vector is distributed.

\begin{lemma}
\label{th:quad_form_own_cov_matrix}
\cite[Chapter 14.6]{Keener2010}
Let $\mathbf{y} \sim \mathcal{N}(\mathbf{y} | \boldsymbol\mu, \mathbf{K})$. Then
\begin{equation}
\mathbf{y}^{T} \mathbf{K}^{-1} \mathbf{y} \sim \chi_n^{' 2} \left( s \right),
\end{equation}
where 
\begin{equation}
s = \sum\limits_{i = 1}^n (\mu'_i)^2, \quad \boldsymbol\mu' = \left[\mu'_i\right]_{i = 1}^n = \mathbf{K}^{-\frac{1}{2}} \boldsymbol\mu.
\end{equation}
\end{lemma} 

\subsection{Quadratic form of an arbitrary symmetric matrix}
Consider the case of a quadratic form of an arbitrary symmetric matrix $\mathbf{K}$ and $\mathbf{y} \sim \mathcal{N}(\mathbf{y} | \boldsymbol\mu, \mathbf{K}')$, where $\mathbf{K} \neq \mathbf{K}'$.

\begin{lemma}
\label{th:arb_quad_form}
\cite{Ruben1962}
Let $\mathbf{y} \sim \mathcal{N}(\mathbf{y} | \boldsymbol\mu, \mathbf{K}')$ and $\mathbf{K}$ be an arbitrary symmetric matrix. Then $\mathbf{y}^{T} \mathbf{K}^{-1} \mathbf{y}$ has a generalised chi-squared distribution.
\end{lemma}
 
\subsection{Proof of Theorem \ref{th:online_distr}}
\label{subsec:stat_distr_th_proof}
\begin{proof}
Recall from~(\ref{eq:online_stat}) the test statistic can be expressed as:
\begin{align}
T_{\text{online}}&(\mathbf{y}_{t - L + 1:t}) \nonumber\\
\label{eq:online_distr}
= & \left(\mathbf{y}_{t-L+1:t} - \boldsymbol\mu_{0}\right)^T \left(\mathbf{K}_{0} + \sigma^2_{0} \mathbf{I} \right)^{-1} \left(\mathbf{y}_{t-L+1:t} - \boldsymbol\mu_{0}\right).
\end{align}

Here $\mathbf{y}_{t-L+1:t} - \boldsymbol\mu_{0} \sim \mathcal{N}(\mathbf{y}_{t-L+1:t} | \mathbf{0}, \mathbf{K}_{0} + \sigma^2_{0} \mathbf{I})$. Therefore, the statistic $T_{\text{online}}(\mathbf{y}_{t - L + 1:t})$ represents a quadratic form of the corresponding covariance matrix. According to Lemma~\ref{th:quad_form_own_cov_matrix}:
\begin{equation}
T_{\text{online}}(\mathbf{y}_{t - L + 1:t}) \sim \chi^{'2}_L(0) = \chi^{2}_L
\end{equation}
where $\chi^{2}_L$ is a chi-squared distribution with $L$ degrees of freedom.
\end{proof}

\subsection{Proof of Theorem \ref{th:power}}
\label{subsec:power_th_proof}
\begin{proof}
Recall that according to the definition (\ref{eq:power_def}) a power of the test is a conditional probability to reject the null hypothesis given that the alternative is true. The hypothesis $\mathcal{H}_0$ is rejected when $T_{\text{online}}(\mathbf{y}_{t-L+1:t}) < e_1$ or $T_{\text{online}}(\mathbf{y}_{t-L+1:t}) > e_2$, where $e_1$ and $e_2$ are the corresponding quantiles, determined in Definition \ref{def:online_test}, and the hypothesis $\mathcal{H}_1$ is true when $\boldsymbol\vartheta = \boldsymbol\vartheta_1$, i.e., $\mathbf{y}_{t-L+1:t} \sim \mathcal{N}(\boldsymbol\mu_{1}, \mathbf{K}_{1} + \sigma_{1}^2 \mathbf{I})$. The power of the test is then:
\begin{align}
B&(\boldsymbol\vartheta) \nonumber\\ 
= &
\mathbb{P}(T_{\text{online}}(\mathbf{y}_{t-L+1:t}) < e_1 \wedge T_{\text{online}}(\mathbf{y}_{t-L+1:t}) > e_2) | \boldsymbol\vartheta = \boldsymbol\vartheta_1) \nonumber\\
= & \mathbb{P}(T_{\text{online}}(\mathbf{y}_{t-L+1:t}) < e_1 | \boldsymbol\vartheta = \boldsymbol\vartheta_1)  \nonumber\\
& + \mathbb{P}(T_{\text{online}}(\mathbf{y}_{t-L+1:t}) > e_2) | \boldsymbol\vartheta = \boldsymbol\vartheta_1)  \nonumber\\
= & \mathbb{P}(T_{\text{online}}(\mathbf{y}_{t-L+1:t}) < e_1 | \boldsymbol\vartheta = \boldsymbol\vartheta_1)  \nonumber\\
& + 1 - \mathbb{P}(T_{\text{online}}(\mathbf{y}_{t-L+1:t}) < e_2 | \boldsymbol\vartheta = \boldsymbol\vartheta_1)  \nonumber\\
= &1 + \mathcal{F}_q(e_1) - \mathcal{F}_q(e_2),
\end{align}
where $\mathcal{F}_q$ is a cdf of a random variable $q$ and:
\begin{equation}
\label{eq:q_in_power}
\begin{split}
q = &T_{\text{online}}(\mathbf{y}_{t-L+1:t}) \\
= & \left(\mathbf{y}_{t-L+1:t} - \boldsymbol\mu_{0}\right)^T (\mathbf{K}_{0} + \sigma_{0}^2 \mathbf{I})^{-1} \left(\mathbf{y}_{t-L+1:t} - \boldsymbol\mu_{0}\right)
\end{split}
\end{equation}
Here $\mathbf{y}_{t-L+1:t}  - \boldsymbol\mu_{0} \sim \mathcal{N}\left(\mathbf{y}_{t-L+1:t} | \boldsymbol\mu_{1} - \boldsymbol\mu_{0}, \mathbf{K}_{1} + \sigma_{1}^2 \mathbf{I}\right)$.

According to Lemma~\ref{th:arb_quad_form}, the quadratic form in (\ref{eq:q_in_power}) has a generalised chi-squared distribution. 

To simplify notation, let us use the following replacements: $\mathbf{y} = \mathbf{y}_{t-L+1:t} - \boldsymbol\mu_{0}$, $\mathbf{K} = \mathbf{K}_{0} + \sigma_{0}^2 \mathbf{I}$, $\mathbf{K}' = \mathbf{K}_{1} + \sigma^2_{1} \mathbf{I}$, and $\boldsymbol\mu = \boldsymbol\mu_{1} - \boldsymbol\mu_{0}$:
\begin{align}
\left(\vphantom{\mathbf{y}_{t-L+1:t}}\right. \mathbf{y}_{t-}&_{L+1:t} - \left.\boldsymbol\mu_{0}\right)^T (\mathbf{K}_{0} + \sigma_{0}^2 \mathbf{I})^{-1} \left(\mathbf{y}_{t-L+1:t} - \boldsymbol\mu_{0}\right) \nonumber\\
= & \mathbf{y}^{T} \mathbf{K}^{-1} \mathbf{y} = \mathbf{y}^{T} {\mathbf{K}'}^{-\frac{1}{2}} {\mathbf{K}'}^{\frac{1}{2}} \mathbf{K}^{-1} {\mathbf{K}'}^{\frac{1}{2}} \underbrace{{\mathbf{K}'}^{-\frac{1}{2}} \mathbf{y}}_{\mathbf{y}'} \nonumber\\
= & {\mathbf{y}'}^{T} {\mathbf{K}'}^{\frac{1}{2}} \mathbf{K}^{-1} {\mathbf{K}'}^{\frac{1}{2}} \mathbf{y}' \nonumber\\
= & \left(\mathbf{y}' - {\mathbf{K}'}^{-\frac{1}{2}} \boldsymbol\mu + {\mathbf{K}'}^{-\frac{1}{2}} \boldsymbol\mu \right)^{T} \nonumber\\
& \times{\mathbf{K}'}^{\frac{1}{2}} \mathbf{K}^{-1} {\mathbf{K}'}^{\frac{1}{2}} \left(\underbrace{\mathbf{y}' - {\mathbf{K}'}^{-\frac{1}{2}} \boldsymbol\mu}_{\mathbf{w}} + {\mathbf{K}'}^{-\frac{1}{2}} \boldsymbol\mu \right) \nonumber\\
\label{eq:arbitrary_quad_form}
\quad= &\left(\mathbf{w} + {\mathbf{K}'}^{-\frac{1}{2}} \boldsymbol\mu \right)^{T} {\mathbf{K}'}^{\frac{1}{2}} \mathbf{K}^{-1} {\mathbf{K}'}^{\frac{1}{2}} \left(\mathbf{w} + {\mathbf{K}'}^{-\frac{1}{2}} \boldsymbol\mu \right).
\end{align}

The matrix ${\mathbf{K}'}^{\frac{1}{2}} \mathbf{K}^{-1} {\mathbf{K}'}^{\frac{1}{2}}$ is symmetric as both matrices $\mathbf{K}$ and $\mathbf{K}'$ are symmetric. Therefore, the matrix ${\mathbf{K}'}^{\frac{1}{2}} \mathbf{K}^{-1} {\mathbf{K}'}^{\frac{1}{2}}$ is normal and the spectral theorem is valid for it:
\begin{equation}
\label{eq:spectral_theorem}
{\mathbf{K}'}^{\frac{1}{2}} \mathbf{K}^{-1} {\mathbf{K}'}^{\frac{1}{2}} = \mathbf{P} \boldsymbol\Lambda \mathbf{P}^T,
\end{equation}
where $\mathbf{P}$ is an orthogonal matrix and $\boldsymbol\Lambda$ is a diagonal matrix. Diagonal elements of $\boldsymbol\Lambda$ are eigenvalues of ${\mathbf{K}'}^{\frac{1}{2}} \mathbf{K}^{-1} {\mathbf{K}'}^{\frac{1}{2}}$ and columns of $\mathbf{P}$ are the corresponding eigenvectors.

Substituting (\ref{eq:spectral_theorem}) into (\ref{eq:arbitrary_quad_form}) the quadratic form is further transformed as:
\begin{align}
\left(\vphantom{{\mathbf{K}'}^{-\frac{1}{2}}}\right.\mathbf{w} + &\left.{\mathbf{K}'}^{-\frac{1}{2}} \boldsymbol\mu \right)^{T} {\mathbf{K}'}^{\frac{1}{2}} \mathbf{K}^{-1} {\mathbf{K}'}^{\frac{1}{2}} \left(\mathbf{w} + {\mathbf{K}'}^{-\frac{1}{2}} \boldsymbol\mu \right) \nonumber\\
\stackrel{\text{eq. (\ref{eq:spectral_theorem})}}{=} & \left(\mathbf{w} + {\mathbf{K}'}^{-\frac{1}{2}} \boldsymbol\mu \right)^{T} \mathbf{P} \boldsymbol\Lambda \mathbf{P}^T \left(\mathbf{w} + {\mathbf{K}'}^{-\frac{1}{2}} \boldsymbol\mu \right) \nonumber\\
= & \left(\mathbf{P}^T \mathbf{w} + \mathbf{P}^T {\mathbf{K}'}^{-\frac{1}{2}} \boldsymbol\mu \right)^{T} \boldsymbol\Lambda \left(\underbrace{\mathbf{P}^T\mathbf{w}}_{\mathbf{w}'} + \underbrace{\mathbf{P}^T {\mathbf{K}'}^{-\frac{1}{2}} \boldsymbol\mu}_{\mathbf{s}} \right) \nonumber\\
\label{eq:arb_quad_form_with_eigen_values}
= & \left(\mathbf{w}' + \mathbf{s} \right)^{T} \boldsymbol\Lambda \left(\mathbf{w}' + \mathbf{s} \right).
\end{align}

As $\boldsymbol\Lambda$ is the diagonal matrix, (\ref{eq:arb_quad_form_with_eigen_values}) can be expressed as:
\begin{equation}
\label{eq:arb_quad_form_final}
\begin{split}
\left(\mathbf{w}' + \mathbf{s} \right)^{T}& \boldsymbol\Lambda \left(\mathbf{w}' + \mathbf{s} \right) =
\sum_{i = 1}^L \lambda_{i} \left(\underbrace{{w'}_i + {s}_i}_{u_i} \right)^2 \\
= & \sum_{i = 1}^L \lambda_{i} \underbrace{u_i^2}_{{v}_i} =
\sum_{i = 1}^L \lambda_{i} {v}_i,
\end{split}
\end{equation}
where $\lambda_{i}$ is the $i$-th diagonal element of the matrix $\boldsymbol\Lambda$, ${w'}_i$ and ${s}_i$ are the $i$-th elements of the vectors $\mathbf{w}'$ and $\mathbf{s}$, respectively, $i \in \{1, \dotsc, L\}$.

Combining (\ref{eq:arbitrary_quad_form}), (\ref{eq:arb_quad_form_with_eigen_values}) and (\ref{eq:arb_quad_form_final}) the input quadratic form can be expressed as:
\begin{equation}
\label{eq:artitraty_quad_form_expr}
\left(\mathbf{y}_{t-L+1:t} - \boldsymbol\mu_{0}\right)^T (\mathbf{K}_{0} + \sigma_{0}^2 \mathbf{I})^{-1} \left(\mathbf{y}_{t-L+1:t} - \boldsymbol\mu_{0}\right) = \sum_{i = 1}^L \lambda_{i} {v}_i.
\end{equation}

Consider now distributions of all the introduced random variables:

\begin{align}
\mathbf{y} &\sim \mathcal{N} (\boldsymbol\mu, \mathbf{K}')\\ 
\mathbf{y}' &= {\mathbf{K}'}^{-\frac{1}{2}} \mathbf{y} &\Rightarrow &\mathbf{y}' \sim \mathcal{N}(\mathbf{y}' | {\mathbf{K}'}^{-\frac{1}{2}} \boldsymbol\mu, \mathbf{I}),\\
\mathbf{w} &= \mathbf{y}' - {\mathbf{K}'}^{-\frac{1}{2}} \boldsymbol\mu &\Rightarrow &\mathbf{w} \sim \mathcal{N}(\mathbf{w} | \mathbf{0}, \mathbf{I}),\\
\mathbf{w}' &= \mathbf{P}^{T} \mathbf{w} &\Rightarrow &\mathbf{w}' \sim \mathcal{N}(\mathbf{w}' | \mathbf{0}, \mathbf{P}^{T} \mathbf{I} \mathbf{P}) = \mathcal{N}(\mathbf{w}' | \mathbf{0}, \mathbf{I}),\\
&\quad &\Rightarrow &{w'}_i \sim \mathcal{N}({w'}_i | 0, 1),\\ 
u_i &= {w'}_i + s_i &\Rightarrow &u_i \sim \mathcal{N}(u_i | {s}_i, 1),\\
\label{eq:chi_sq_terms}
{v}_i &= u_i^2 &\Rightarrow &{v}_i \sim \chi_1^{' 2} (s_i^2)
\end{align}

Substituting back the values for $\mathbf{K}$, $\mathbf{K}'$, and $\boldsymbol{\mu}$ we get the theorem statement. 
\end{proof}

\section{Performance validation on synthetic data}
\label{sec:synthetic}

\begin{figure}[!t]
\centering
\includegraphics[width = 0.35\textwidth]{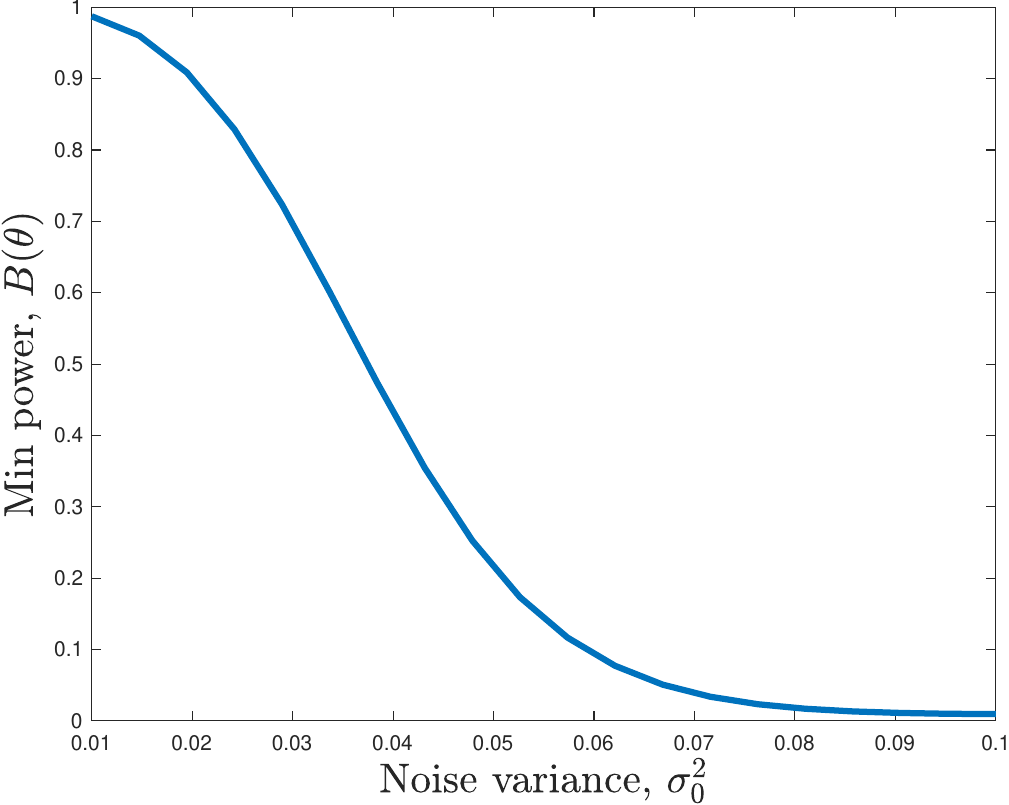}
\caption{Minimum power of the GP-OLCDT within the area $A$ of hyperparameters for the alternative $\mathcal{H}_1$ as a function of the variance $\sigma^2_0$ of the additive noise.}
\label{fig:power_noise}
\end{figure}

This section presents a numerical evaluation of the proposed methods for change point detection on synthetic data. 

The following performance measures are used~\cite{gustafsson2000adaptive, Basseville1993}:

\begin{itemize}
\item \textit{Mean time between false alarms}: $\mathbb{E}(t_a - t_0 | \text{ no change})$, where $\mathbb{E}$ is the mathematical expectation, $t_a$ is a time of an alarm, generated by a change point detection algorithm, and this alarm is false, $t_0$ is a starting time. We consider the previous false alarm or a true missed change point as a starting time;
\item \textit{Mean delay for detection}: $\mathbb{E}(t_a - t^{*})$, where $t^{*}$ is a true time of a change;
\item \textit{Missed detection rate}: a probability of not receiving an alarm, when there has been a change, which can be estimated as $\dfrac{\#\{\text{missed detections}\}}{\#\{\text{change points}\}}$. 
\end{itemize}

\begin{figure*}[!ht]
\centering
\subfloat[$\sigma^2_0 = 0.03$]{\includegraphics[width = 0.9\columnwidth]{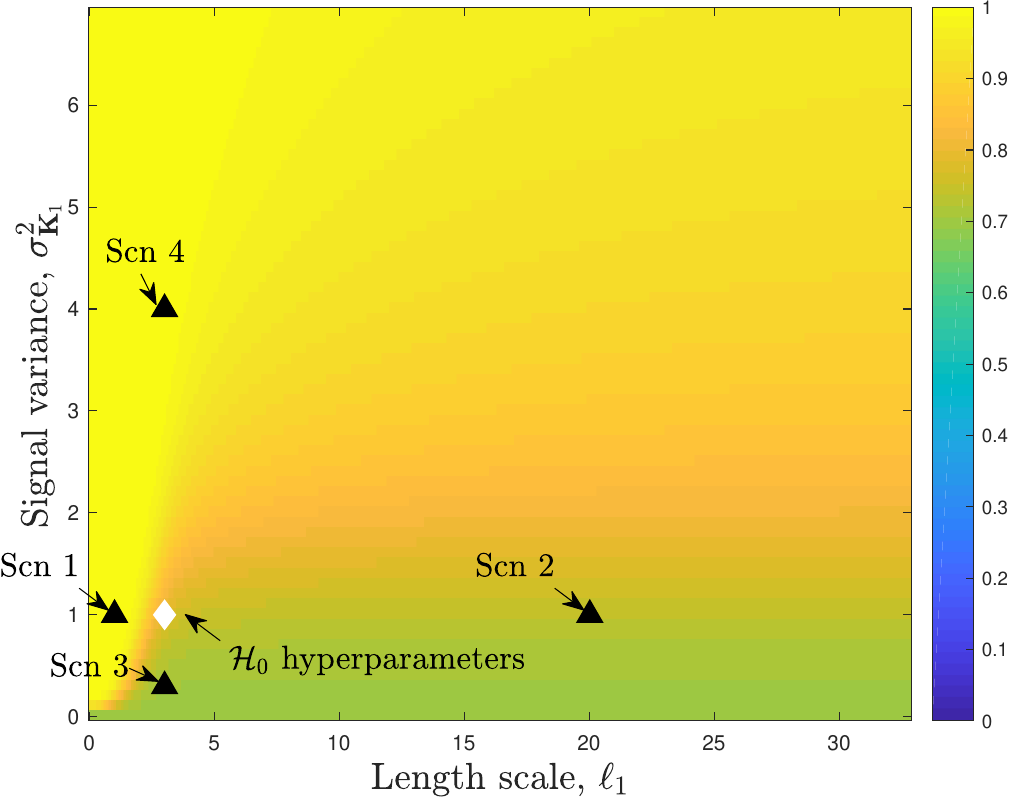}}
\hfil
\subfloat[$\sigma^2_0 = 0.1$]{\includegraphics[width = 0.9\columnwidth]{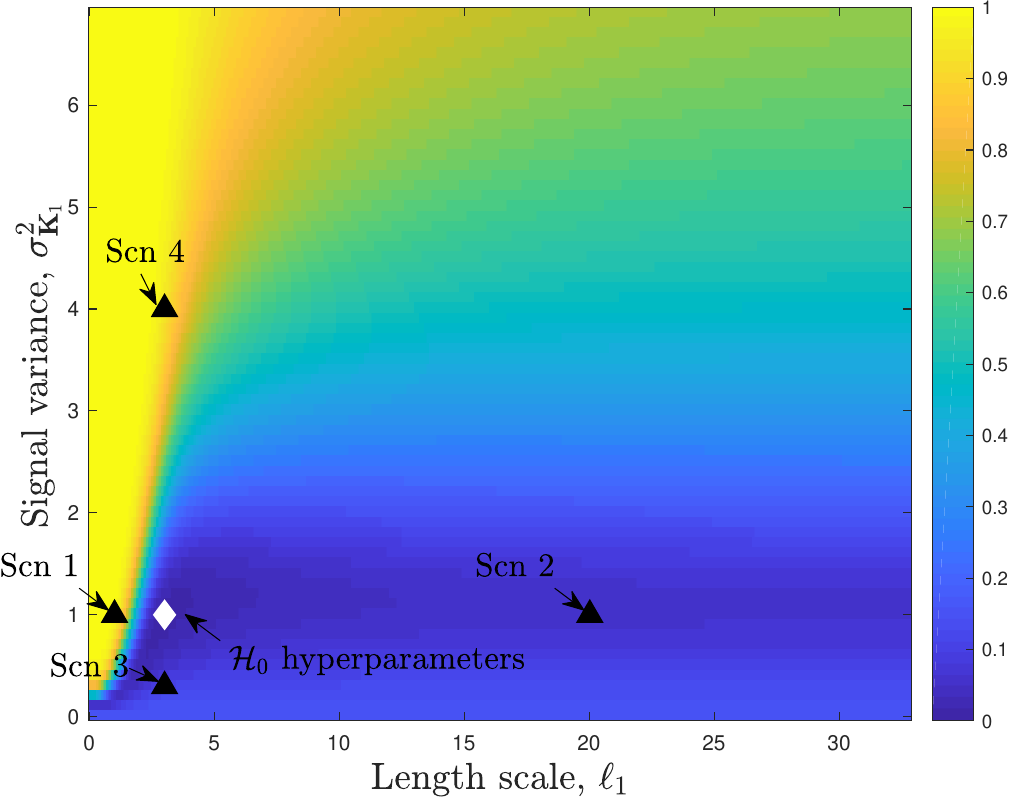}
\label{fig:power_fun_0.1}}
\caption{Power of the GP-OLCDT as a function of the hyperparameters for the alternative $\mathcal{H}_1$. The null hypothesis hyperparameters are plotted as a diamond. Triangular points correspond to the values of the alternative hyperparameters from the four considered scenarios.}
\label{fig:power_fun}
\end{figure*}

A change point detection algorithm is required to maximise the first measure and minimise the other two. A reliable algorithm should achieve a tradeoff with respect to the three criteria. Increase of the number of algorithm detections leads to reduction of the missed detection rate, although there is a risk to increase number of false alarms and hence it might negatively affect the mean time between false alarms measure. On the other hand, reduction of the number of detections might improve the performance in terms of false alarms, however, it might also increase the missed detection rate.

It is a well-known problem~\cite{gustafsson2000adaptive, Basseville1993}, which of detections to treat as correct detections while considering the others as false alarms. For example, an alarm at time point $t_a > t^{*}$, where $t^{*}$ is a true time of a change, can be considered as a delayed correct detection or as a false alarm. In this paper when the detection comes within a time window around the true change point it is considered as a correct detection. The time difference between this detection and the true change point contributes to the mean delay for detection performance measure. If the detection comes outside of this time window it is treated as a false alarm. Therefore, the mean time between false alarms is updated.    

\subsection{Data simulated by the proposed generative model}
\label{sec:our_syn_data}

A time series generated by the model (\ref{eq:change_data_model}) is considered, where the change point time $t^{*}$ is set to $101$. The total number $N$ of observed points is equal to $200$. Observed data points are acquired at time moments $\mathbf{t}_{1:N} = 1, 2, \dotsc, 200$. Function values are sampled based on the GP: $\mathbf{f}_{1:100} \sim \mathcal{N}\left(\mathbf{f}_{1:100} | \mathbf{0}, \mathbf{K}_0\right)$ and $\mathbf{f}_{101:200} \sim \mathcal{N}\left(\mathbf{f}_{101:200} | \mathbf{0}, \mathbf{K}_1\right)$ before and after the change, respectively. Squared exponential covariance matrices are used: 
\begin{equation}
\mathbf{K}_{a} (i, j) = \sigma_{\mathbf{K}_{a}}^2 \exp\left(-\dfrac{(t_i - t_j)^2}{2 \ell_{a}^2}\right), \quad a \in \{0, 1\},
\end{equation}
with parameters: a signal variance $\sigma^2_{\mathbf{K}_{a}}$ and a length scale~$\ell_{a}$. 

Corresponding noisy observations $\mathbf{y}_{1:100}$ and $\mathbf{y}_{101:200}$ are obtained by adding the Gaussian distributed noise, which variance is set to $\sigma^2$ both before and after the change.

\begin{figure}[!t]
\centering
\subfloat[Scenario 1]{\includegraphics[width = 0.4\columnwidth]{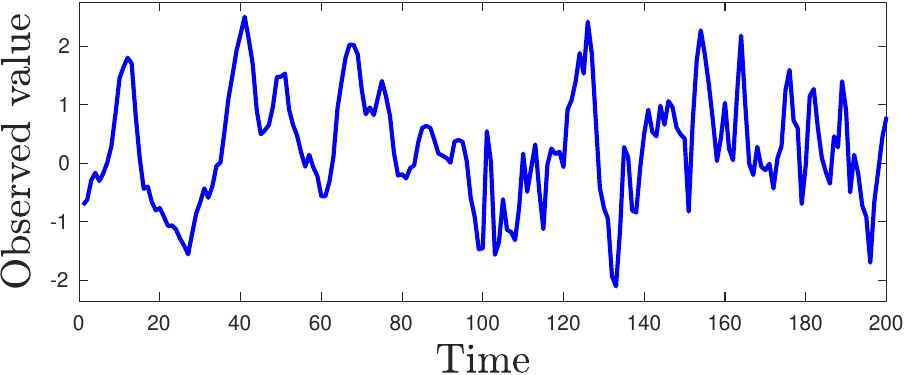}}
\hfil
\subfloat[Scenario 2]{\includegraphics[width = 0.4\columnwidth]{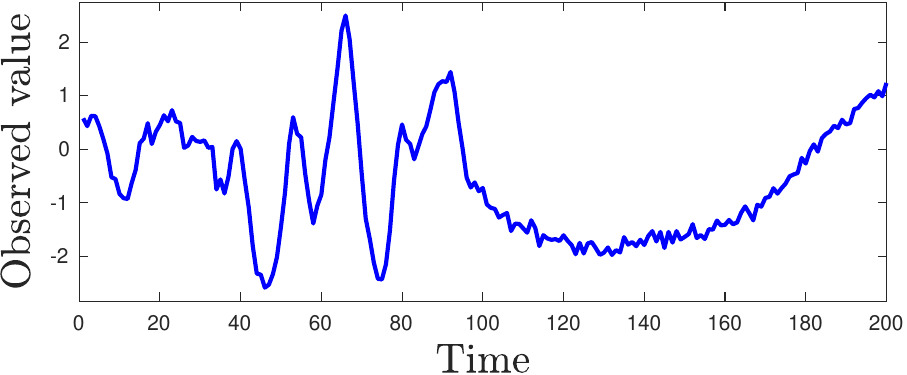}}
\hfil
\subfloat[Scenario 3]{\includegraphics[width = 0.4\columnwidth]{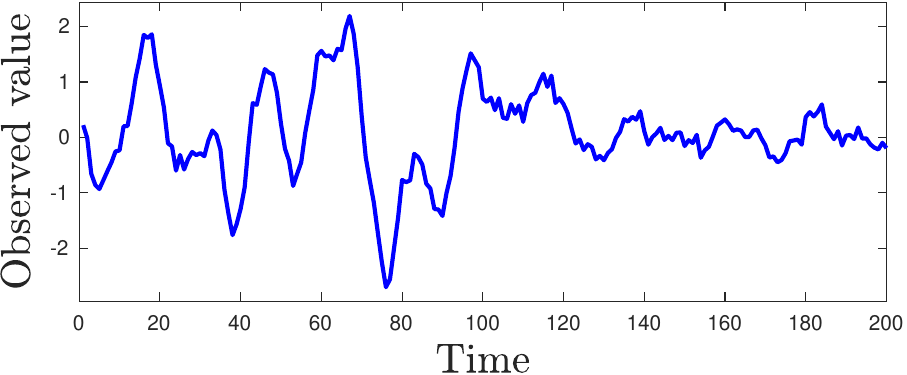}}
\hfil
\subfloat[Scenario 4]{\includegraphics[width = 0.4\columnwidth]{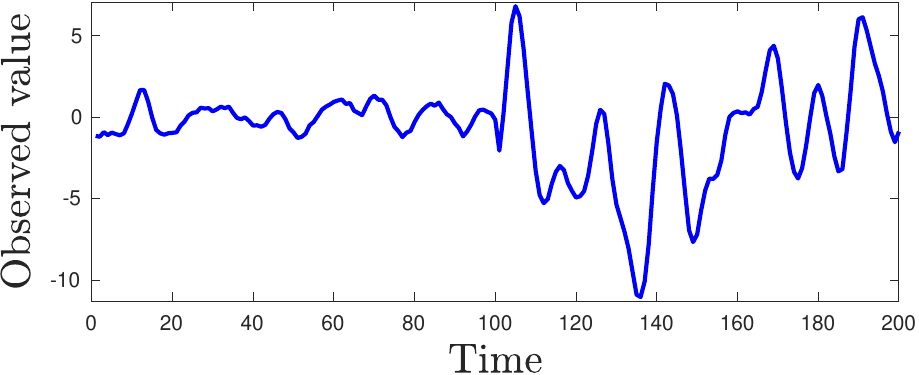}}
\caption{Sample observations generated under four scenarios by the proposed generative model.}
\label{fig:our_data_example}
\end{figure} 

\begin{table}[!t]
  \caption{Scenarios characteristics for data simulated by the proposed generative model. The null hypothesis hyperparameters are: $\ell_0 = 3$ and $\sigma^2_{\mathbf{K}_0} = 1$}
  \label{tab:power}
  \centering
  \begin{tabular}{p{1.1cm} p{1.3cm} p{1.5cm} p{2.9cm}}
    \toprule
    Scenario     & Length scale, $\ell_1$ & Signal variance, $\sigma^2_{\mathbf{K}_1}$ & Power of the GP-OLCDT with known hyperparameters\\
    \midrule
    1 & 1 & 1 & 0.99\\
	2 & 20 & 1 & 0.06\\
	3 & 3 & 0.3 & 0.10\\
	4 & 3 & 4 & 0.95\\
    \bottomrule
  \end{tabular}
\end{table} 

\begin{figure*}[!t]
\centering
\subfloat{
	\scalebox{1}{\begin{tikzpicture}
\begin{axis}[
  hide axis,
  xmin=0,
  xmax=1,
  ymin=0,
  ymax=1,
  legend columns=-1,
  legend style={draw=white!15!black},
  legend style={/tikz/every even column/.append style={column sep=0.5cm}},
  legend cell align={left}
]
\addlegendimage{solid,color={rgb,1:red,0; green,0; blue,1}, line width=1.5pt}
\addlegendentry{\scriptsize{given $\mathcal{H}_0$ is true}}
\addlegendimage{dashed,color={rgb,1:red,1; green,0; blue,0}, line width=1.5pt}
\addlegendentry{\scriptsize{given $\mathcal{H}_1$ is true}}
\addlegendimage{solid,color={rgb,1:red,0; green,0; blue,0}, line width=1.5pt}
\addlegendentry{\scriptsize{$\frac{\alpha}{2}$ \& $1-\frac{\alpha}{2}$ quantiles}}
\addlegendimage{only marks,mark=x,mark options={mark size=3pt, line width = 1pt}, color={rgb,1:red,1; green,0; blue,1}}
\addlegendentry{\scriptsize{power \vphantom{$\mathcal{H}_0$}}}
\end{axis}
\end{tikzpicture}}
}
\newline
\subfloat[Scenario 1]{\includegraphics[width=0.4\textwidth]{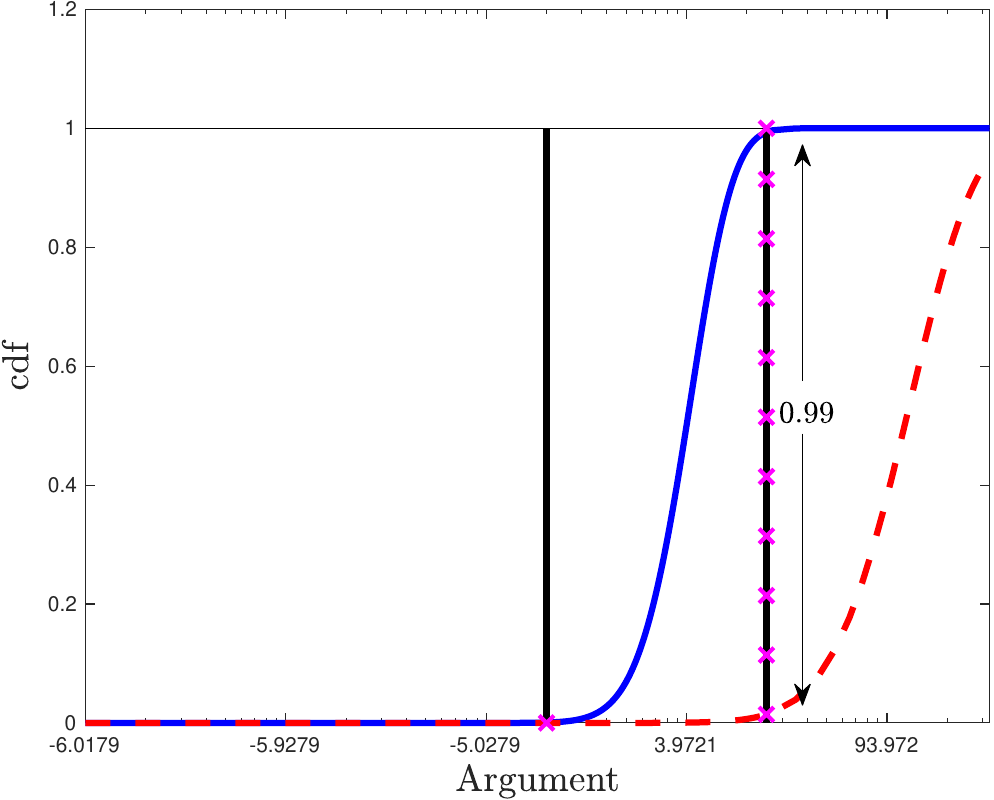}}
\hfil
\subfloat[Scenario 2]{\includegraphics[width=0.4\textwidth]{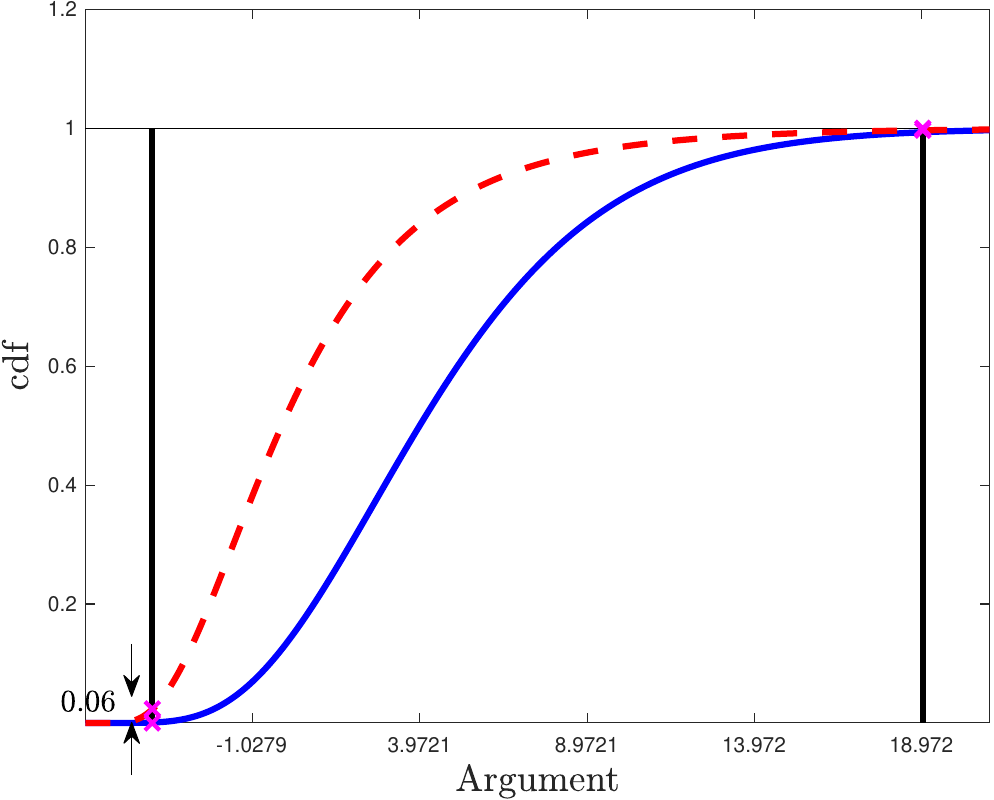}}
\hfil
\subfloat[Scenario 3]{\includegraphics[width=0.4\textwidth]{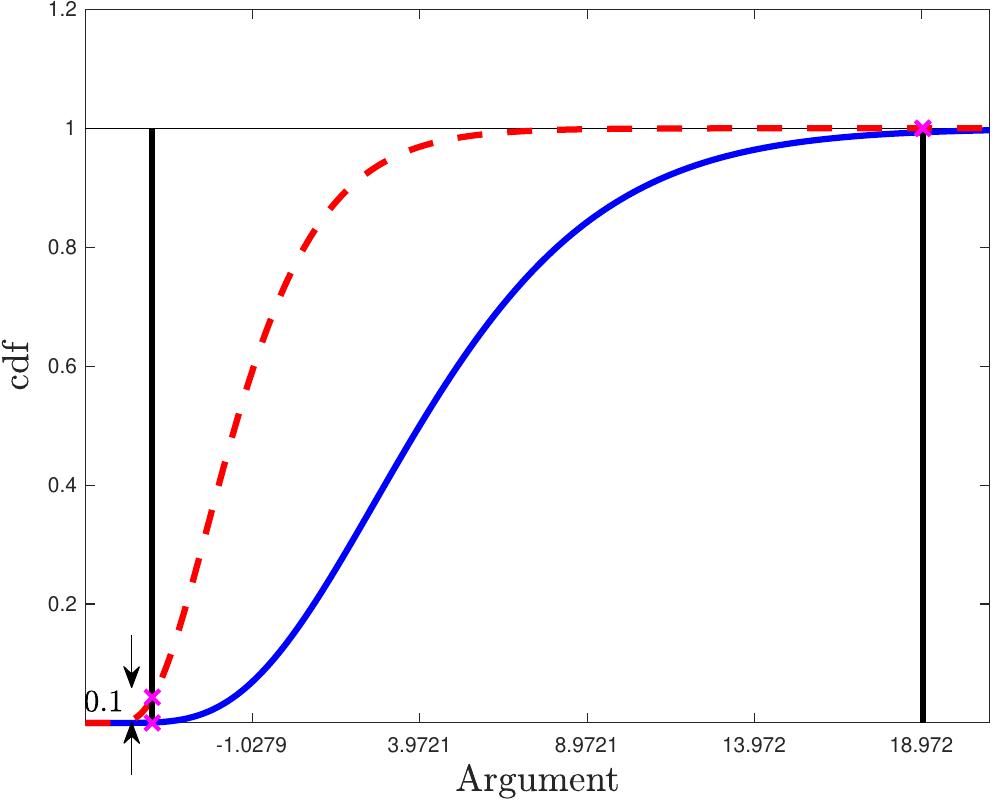}}
\hfil
\subfloat[Scenario 4]{\includegraphics[width=0.4\textwidth]{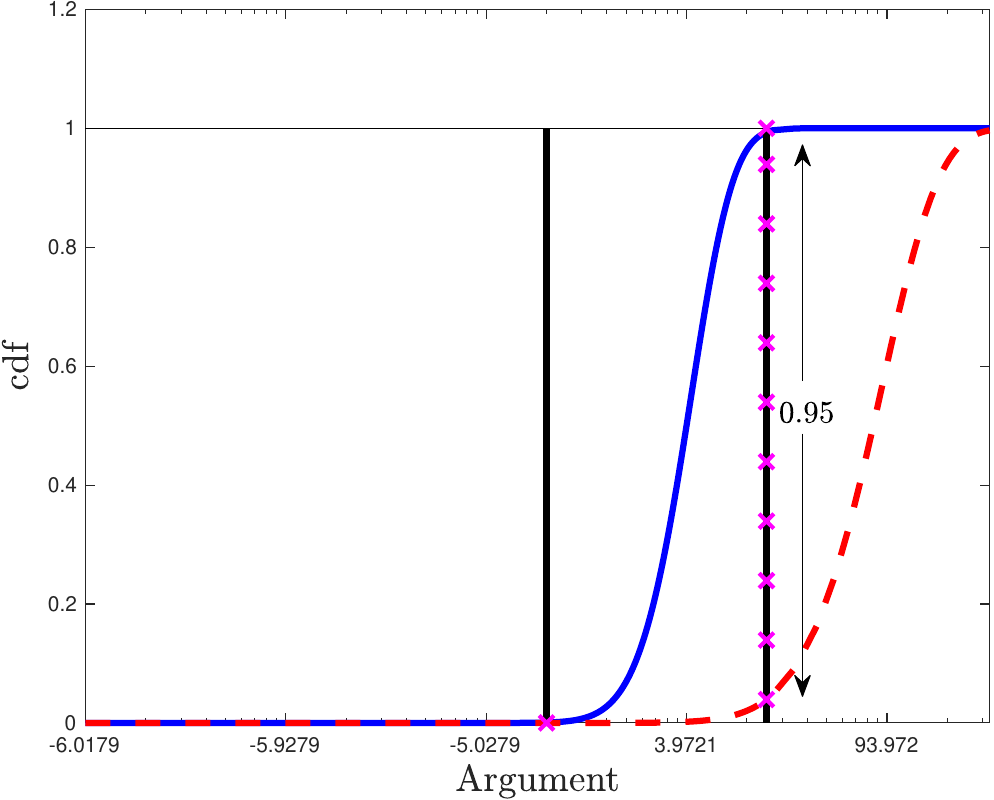}}
\caption{Cdfs of the test statistic for the GP-OLCDT with the known hyperparameter vector $\boldsymbol\vartheta$ under the $\mathcal{H}_0$ and $\mathcal{H}_1$ hypotheses. The vertical lines correspond to $\alpha_{\text{stat}} / 2$ and $1 - \alpha_{\text{stat}} / 2$ quantiles of the the test statistic distribution under the null hypothesis $\mathcal{H}_0$, the significance level $\alpha_{\text{stat}}$ is set to $0.01$. These quantiles are used as lower and upper bounds for the test statistic to make a decision of rejecting $\mathcal{H}_0$. The power of the test values are also marked in the plots.}
\label{fig:statistic_disctr}
\end{figure*}

Theorem~\ref{th:power} allows to compute the power of the GP Online Likelihood-based Change point Detection Test (GP-OLCDT) for any given hyperparameter vectors~$\boldsymbol\vartheta_0$ and~$\boldsymbol\vartheta_1$. In the considered settings only the covariance function hyperparameters are changing:~$\boldsymbol\vartheta_a = \{\sigma^2_{\mathbf{K}_a}, \ell_a\}$,~$a \in \{0, 1\}$. We first evaluate how the power of the GP-OLCDT depends on the additive noise variance $\sigma^2_0$. The hyperparameters~$\boldsymbol\vartheta_0$ are fixed as:~$\ell_0 = 3$ and $\sigma^2_{\mathbf{K}_0} = 1$. The hyperparameters $\boldsymbol\vartheta_1$ vary within the area $A$: $\{\ell_1 = (0, 33) \text{ and } \sigma^2_{\mathbf{K}_0} = (0, 7)\}$. The additive noise variance $\sigma^2_1 = 0.1$ for the alternative is fixed. Figure~\ref{fig:power_noise} shows how the minimum power within the area $A$ depends on the additive noise variance $\sigma^2_0$ for the null hypothesis. For the small level of noise $\sigma^2_0 = 0.01$ the power of the test is equal to 1 within the whole area $A$, this means that for any alternative which hyperparameters lie within this area the GP-OLCDT would reject the null hypothesis when the alternative is true with the probability $1$. For the larger level of noise $\sigma^2_0 = 0.1$ there are such values of hyperparameters~$\boldsymbol\vartheta_1$ that the power of the GP-OLCDT is almost $0$.  

Figure~\ref{fig:power_fun} presents the power of the test values for the area $A$ for the noise level $\sigma^2_0 = 0.03$ and $\sigma^2_0 = 0.1$. In the current settings the proposed test performs better in the regions where the length scale is less after the change than before it and the signal variance changes from small to large values before and after the change, respectively. For the later analysis we only consider the more challenging settings with the noise level $\sigma^2_0 = \sigma^2_1 = 0.1$.

The following scenarios are examined in more details:
\begin{enumerate}
\item $\ell_0 > \ell_1$ and $\sigma^2_{\mathbf{K}_0} = \sigma^2_{\mathbf{K}_1}$;
\item $\ell_0 < \ell_1$ and $\sigma^2_{\mathbf{K}_0} = \sigma^2_{\mathbf{K}_1}$;
\item $\ell_0 = \ell_1$ and $\sigma^2_{\mathbf{K}_0} > \sigma^2_{\mathbf{K}_1}$;
\item $\ell_0 = \ell_1$ and $\sigma^2_{\mathbf{K}_0} < \sigma^2_{\mathbf{K}_1}$.
\end{enumerate}

The value of the hyperparameter vector $\boldsymbol\vartheta_0$ is fixed for all four scenarios: $\ell_0 = 3$ and $\sigma^2_{\mathbf{K}_0} = 1$. The values for the hyperparameter vector $\boldsymbol\vartheta_1$ after the change are provided in Table~\ref{tab:power}. Table~\ref{tab:power} also presents the power of the GP-OLCDT for each scenario. The example of data generated for each of the scenarios can be found in Figure~\ref{fig:our_data_example}.

\begin{table*}[!t]
  \caption{Change point detection performance on data simulated by the proposed generative model}
  \label{tab:mean_results_our_data}
  \centering
  \begin{tabular}{p{5.2cm} p{1.5cm} p{1.5cm} p{1.2cm} p{1.5cm}}
    \toprule
    Method     &  Mean time between false alarms  & Mean delay for detection & Missed detection rate & Power\\
    \midrule
    \textbf{Scenario 1} & & & &\\
    GP-OLCDT with known $\boldsymbol\vartheta_0$  & $157.67$ & $1.31$ & $\mathbf{0.00}$ & $0.99$\\
GP-OLCDT with batch estimated $\boldsymbol\vartheta_0$  & $164.53$ & $1.24$ & $\mathbf{0.00}$ & $0.99\pm0.00$\\
GP-OLCDT with sliding estimated $\boldsymbol\vartheta_0$  & $48.19$ & $1.56$ & $0.01$ & $0.99\pm0.01$\\
GP-OLCDT with adaptively estimated $\boldsymbol\vartheta_0$ & $87.54$ & $6.48$ & $0.08$ & $0.99\pm0.01$\\
GP-LRT & $\mathbf{195.95}$ & $\mathbf{0.60}$ & $0.04$ &  ---\\
GP-GLRT & $177.75$ & $0.65$ & $0.19$ &  ---\\
GP-BOCPD & $129.65$ & $2.40$ & $\mathbf{0.00}$ &  ---\\
    \midrule
    \textbf{Scenario 2} & & & &\\
GP-OLCDT with known $\boldsymbol\vartheta_0$  & $157.67$ & $11.50$ & $0.10$ & $0.06$\\
GP-OLCDT with batch estimated $\boldsymbol\vartheta_0$  & $164.53$ & $10.83$ & $0.13$ & $0.06\pm0.02$\\
GP-OLCDT with sliding estimated $\boldsymbol\vartheta_0$  & $86.18$ & $8.82$ & $0.23$ & $0.11\pm0.04$\\
GP-OLCDT with adaptively estimated $\boldsymbol\vartheta_0$ & $75.35$ & $15.26$ & $0.15$ & $0.11\pm0.08$\\
GP-LRT & $190.88$ & $\mathbf{0.15}$ & $0.09$ &  ---\\
GP-GLRT & $\mathbf{192.14}$ & $25.80$ & $\mathbf{0.07}$ &  ---\\
GP-BOCPD & $190.56$ & $11.52$ & $0.37$ &  ---\\
    \midrule
    \textbf{Scenario 3} & & & &\\
GP-OLCDT with known $\boldsymbol\vartheta_0$  & $157.67$ & $13.63$ & $\mathbf{0.02}$ & $0.10$\\
GP-OLCDT with batch estimated $\boldsymbol\vartheta_0$  & $164.53$ & $14.16$ & $0.08$ & $0.10\pm0.04$\\
GP-OLCDT with sliding estimated $\boldsymbol\vartheta_0$  & $108.36$ & $17.11$ & $0.30$ & $0.09\pm0.04$\\
GP-OLCDT with adaptively estimated $\boldsymbol\vartheta_0$ & $86.77$ & $19.41$ & $0.17$ & $0.08\pm0.07$\\
GP-LRT & $185.57$ & $\mathbf{0.67}$ & $0.14$ &  ---\\
GP-GLRT & $183.02$ & $14.20$ & $0.15$ &  ---\\
GP-BOCPD & $\mathbf{195.65}$ & $13.28$ & $0.50$ &  ---\\
    \midrule
    \textbf{Scenario 4} & & & &\\
GP-OLCDT with known $\boldsymbol\vartheta_0$  & $157.67$ & $0.64$ & $\mathbf{0.00}$ & $0.95$\\
GP-OLCDT with batch estimated $\boldsymbol\vartheta_0$  & $164.53$ & $0.56$ & $\mathbf{0.00}$ & $0.96\pm0.02$\\
GP-OLCDT with sliding estimated $\boldsymbol\vartheta_0$  & $68.23$ & $0.61$ & $0.03$ & $0.96\pm0.02$\\
GP-OLCDT with adaptively estimated $\boldsymbol\vartheta_0$ & $70.76$ & $8.70$ & $0.04$ & $0.97\pm0.03$\\
GP-LRT & $\mathbf{196.91}$ & $\mathbf{0.30}$ & $0.03$ &  ---\\
GP-GLRT & $189.99$ & $0.82$ & $0.08$ &  ---\\
GP-BOCPD & $188.13$ & $2.14$ & $\mathbf{0.00}$ &  ---\\
    \bottomrule
  \end{tabular}
\end{table*}

Based on the results given in Figure~\ref{fig:power_fun_0.1} and Table~\ref{tab:power} the proposed GP-OLCDT is expected to perform better under the $1$-st and $4$-th scenarios while having difficulties under the $2$-nd and $3$-rd scenarios. Indeed, in the $2$-nd and $3$-rd scenarios the hyperparameter vector~$\boldsymbol\vartheta_0$ value before a change represents more flexible settings that allow to explain larger variety of data. Therefore, the data after the change, based on the hyperparameter vector $\boldsymbol\vartheta_1$, can be also fitted with the model before the change. The likelihood function value would not then drop significantly after the change, which causes low performance of the test.

The distribution of the test statistic for the GP-OLCDT can be determined based on Theorem~\ref{th:online_distr}. The cdfs of the test statistic under the null hypothesis $\mathcal{H}_0$ and alternative $\mathcal{H}_1$ for all four scenarios are presented in Figure~\ref{fig:statistic_disctr}. The figure demonstrates that the cdfs of the test statistic under the null hypothesis and alternative are very close to each other for the $2$-nd and $3$-rd scenarios and it is difficult to distinguish the two cdfs. It further explains the reasons of low power of the test values for these two scenarios.

For each of the scenarios the following methods are applied: 
\begin{itemize}
\item the GP-OLCDT with known $\boldsymbol\vartheta_0$ (Section~\ref{subsec:online_test_known});
\item the GP-OLCDT with batch estimated $\hat{\boldsymbol\vartheta}_0$, where the estimate is obtained by optimising marginal likelihood on the first part of data $\mathbf{y}_{1:100}$ (Section~\ref{subsec:online_test_known});
\item the GP-OLCDT with sliding estimated $\hat{\boldsymbol\vartheta}_0$, where the estimate is obtained by optimising marginal likelihood on data $\mathbf{y}_{t - \tilde{L} : t-1}$ from a sliding window (Section~\ref{subsubsec:sliding_window_opt}), $\tilde{L} = 25$;
\item the GP-OLCDT with adaptively estimated $\hat{\boldsymbol\vartheta}_0$, where the estimate is obtained by optimising marginal likelihood on the first $\tilde{L}$ data points after each detected change (Section~\ref{subsubsec:adaptive_opt}), $\tilde{L} = 25$;
\item the GP log likelihood ratio change point detection test (GP-LRT) with known $\boldsymbol\vartheta_0$ and $\boldsymbol\vartheta_1$ (Section \ref{subsec:lrt});
\item the GP generalised log likelihood ratio change point detection test (GP-GLRT) with estimated $\hat{\boldsymbol\vartheta}_0$ and $\hat{\boldsymbol\vartheta}_1$, where the estimates are obtained based on first and last $M = 25$ data points, respectively (Section \ref{subsec:gen_lrt}).
\end{itemize}
For all the GP-OLCDTs the significance level is set to $\alpha_{\text{stat}} = 0.01$ and sliding window width is set to $L = 10$.  

The proposed tests are also compared with the stationary GP Bayesian Online Change Point Detection (GP-BOCPD) algorithm~\cite{Saatci2010}. In the stationary GP-BOCPD method the data~$y_t$ is also assumed to be generated by a GP time series model. In contrast to the proposed framework, the functions~$f_0$ and~$f_1$ before and after a change are assumed to be different realisations of the same GP prior, whereas in the proposed generative model the hyperparameters of the GP are different before and after the change.

The mean function in GP-BOCPD is also set to zero. The covariance function is the sum of a squared exponential, a periodic, a constant, and ``white noise'' covariance functions~\cite{rasmussen2006gaussian}. The first $100$ data points are used for hyperparameter learning. Within the GP-BOCPD framework change points are estimated via a posterior probability of a run-length, i.e., the length of a time period since the last change point. An alarm is generated if the posterior probability of a change point at a considered time moment is more than a threshold. Here the threshold is set to $0.99$.

%For each of the four scenarios the comparing methods are first examined based on one random realisation of observed data $\mathbf{y}$ and then average results of the performance measures are discussed based on $100$ Monte Carlo runs.

%Recall Theorem~\ref{th:power}, a power of a test depends on hyperparameters both before and after a change. Here the hyperparameter values actually used in a test are used as $\boldsymbol\vartheta_0$, i.e., for the GP-OLCDT with known hyperparameters this is the true $\boldsymbol\vartheta_0$, for both batch and sliding estimated hyperparameters, this is the obtained estimates $\hat{\boldsymbol\vartheta}_0$. For all the tests the true values $\boldsymbol\vartheta_1$ are used for hyperparameters after a change in power computation. For the GP-OLCDT with sliding estimated $\hat{\boldsymbol\vartheta}_0$ the power then varies as new data is processed. Figure~\ref{fig:power} presents the power values for the GP-OLCDT with sliding estimated $\hat{\boldsymbol\vartheta}_0$ for each time moment. The estimates $\hat{\boldsymbol\vartheta}_0$ of the hyperparameter values before a change for both the GP-OLCDT using estimated hyperparameters 

Consider the average performance of all the methods based on $100$ Monte Carlo simulations for each of the scenarios. Table~\ref{tab:mean_results_our_data} presents the obtained values for the performance measures. Here the first alarm generated after the true change time is treated as a correct detection, whereas other alarms are processed as false alarms. 

The empirical performance of the GP-OLCDTs in terms of the missed detection rate is superior than the theoretical power estimates for the $2$-nd and $3$-rd scenarios (the missed detection rate can be considered as a complement to the power of a test). In the $1$-st and $4$-th scenarios the empirical missed detection rate results of the GP-OLCDTs confirm the theoretical power performance. The analysis of the results given in Table~\ref{tab:mean_results_our_data} shows though that the $2$-nd and $3$-rd scenarios are much more difficult for the analysis for all the methods in comparison to the other two scenarios. 

\begin{figure}[!t]
\centering
\subfloat[``Easy'' scenario]{\includegraphics[width = 0.45\columnwidth]{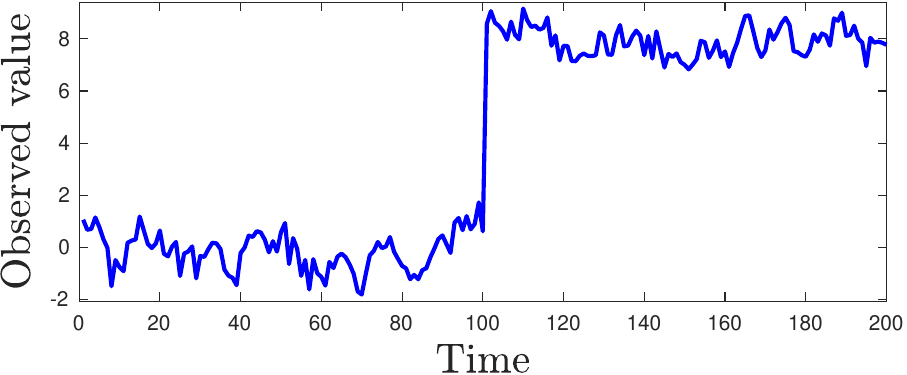}}
\hfil
\subfloat[``Difficult'' scenario]{\includegraphics[width = 0.45\columnwidth]{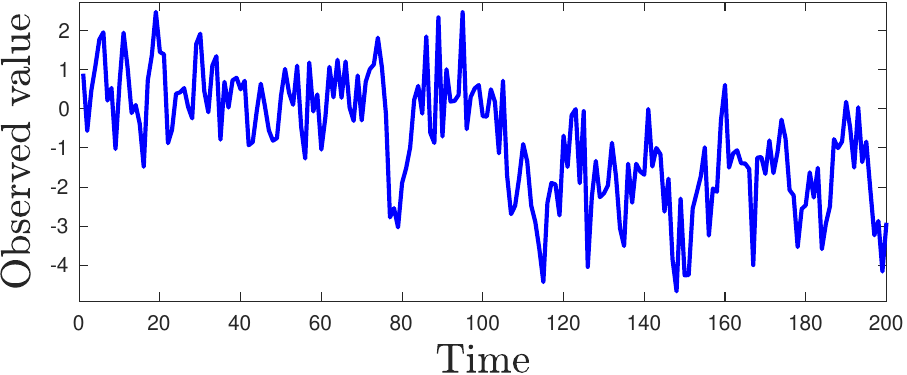}}
\caption{Sample observations generated under 2 scenarios by the GP-BOCPD generative model.}
\label{fig:bocpd_data_example}
\end{figure}

\begin{table}[!t]
  \caption{Change point detection performance on data simulated by the GP-BOCPD model}
  \label{tab:mean_results_bocpd_data}
  \centering
  \begin{tabular}{p{2.9cm} p{1.5cm} p{1.3cm} p{1.2cm}}
    \toprule
    Method     &  Mean time between false alarms  & Mean delay for detection & Missed detection rate\\
    \midrule
    \textbf{``Easy'' scenario} & & &\\
GP-OLCDT with batch estimated $\boldsymbol\vartheta_0$  & $151.37$ & $6.59$ & $\mathbf{0.05}$\\
GP-OLCDT with sliding estimated $\boldsymbol\vartheta_0$  & $94.48$ & $4.25$ & $0.24$\\
GP-OLCDT with adaptively estimated $\boldsymbol\vartheta_0$ & $71.86$ & $9.42$ & $0.16$\\
GP-GLRT & $181.19$ & $\mathbf{0.87}$ & $0.30$\\
GP-BOCPD & $\mathbf{197.85}$ & $10.13$ & $0.25$\\
    \midrule
    \textbf{``Difficult'' scenario} & & &\\
GP-OLCDT with batch estimated $\boldsymbol\vartheta_0$  & $158.93$ & $\mathbf{15.34}$ & $\mathbf{0.17}$\\
GP-OLCDT with sliding estimated $\boldsymbol\vartheta_0$  & $144.66$ & $20.24$ & $0.54$\\
GP-OLCDT with adaptively estimated $\boldsymbol\vartheta_0$ & $92.19$ & $18.58$ & $0.23$\\
GP-GLRT & $154.06$ & $20.63$ & $0.35$\\
GP-BOCPD & $\mathbf{191.49}$ & $32.73$ & $0.60$\\
    \bottomrule
  \end{tabular}
\end{table}

The GP-OLCDTs demonstrate better results than the GP-BOCPD in terms of the delay for detection and comparable outputs with respect to both the time between false alarms and the missed detection rate for the $1$-st and $4$-th scenarios. In the $2$-nd and $3$-rd scenarios the GP-OLCDTs outperform the GP-BOCPD in terms of the missed detection rate having the similar results for both the time between false alarms and the delay for detection. However, the GP-OLCDT with sliding estimated $\boldsymbol\vartheta_0$ has a slightly larger number of false alarms causing lower values of the corresponding performance measure than the other methods, and the GP-OLCDT with adaptively estimated $\boldsymbol\vartheta_0$ has slightly more delayed detections than the other methods. Likelihood ratio tests are able to find the most accurate estimates of change times (the smallest mean delay for detection) in all the cases. The GP-LRT and GP-GLRT has higher results in terms of missed detection rate because they can only predict one change point for a sequence and if it is predicted before the actual change point it is counted as missed detection.   

\subsection{Data simulated by the GP-BOCPD model}
To make a fair comparison the methods are also applied to data generated by the GP-BOCPD model, i.e., $\mathbf{f}_{1:100}$ and $\mathbf{f}_{101:200}$ are different realisations of the same GP with fixed hyperparameters. 

For data generation the same settings for the mean and covariance functions are used as in the GP-BOCPD algorithm in Section~\ref{sec:our_syn_data}. 

Although the proposed methods are not limited to any particular type of covariance functions, the squared exponential covariance function is used. As the data is generated based on the other model rather than the one used in the test assumptions the GP-OLCDT with known hyperparameters and GP-LRT are not applicable in this setting. 

Two scenarios are considered: (1) an ``easy'' one, where the changes are well distinguishable, and (2) a ``difficult'' one, where the changes are not easily detected. Figure~\ref{fig:bocpd_data_example} presents sample data obtained under both scenarios.

Average performance results among $100$ Monte Carlo runs can be found in Table~\ref{tab:mean_results_bocpd_data}. Although the proposed methods use the data model assumptions different to the actual generative model, they demonstrate better results than the GP-BOCPD in terms of both the delay for detection and the missed detection rate for the two scenarios. However, the proposed methods generate more false alarms than the GP-BOCPD that leads to the lower performance in terms of the time between false alarms. 

\begin{figure}[!t]
\centering
\subfloat[Sample frame]{\includegraphics[width=0.47\columnwidth]{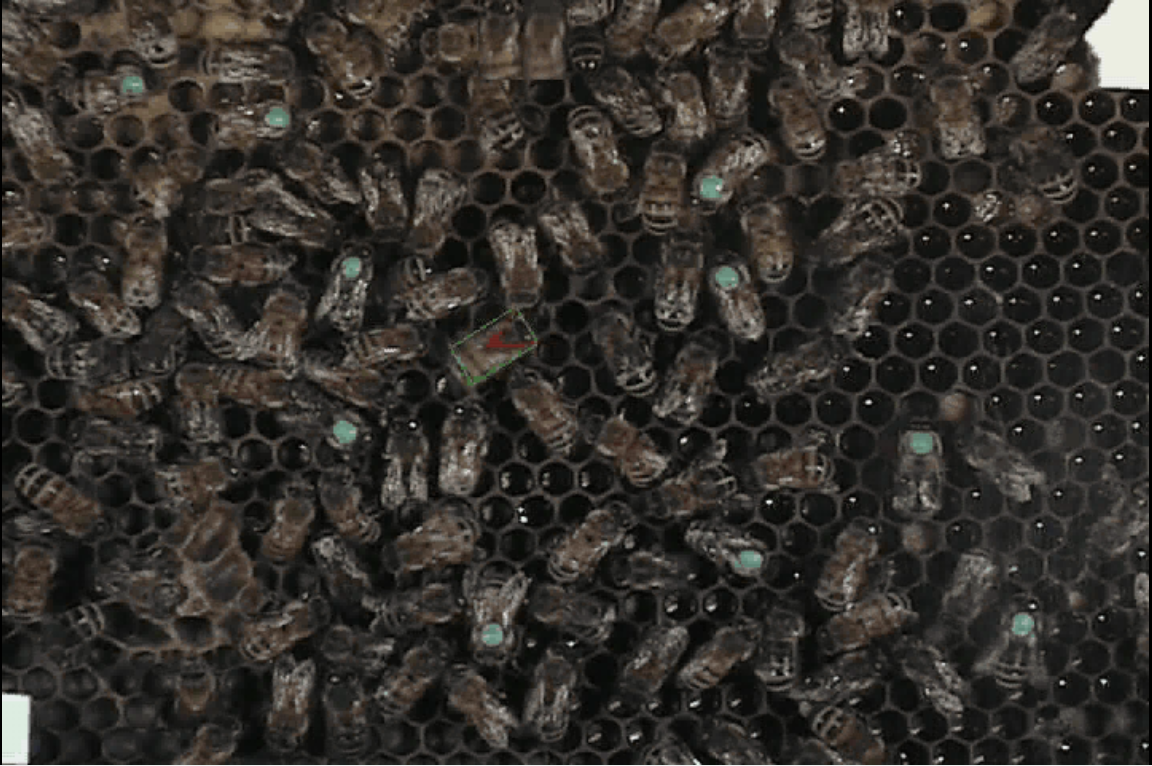}
\label{fig:bee_data_sample}}
\hfil
\subfloat[Dance phases]{\includegraphics[width=0.47\columnwidth]{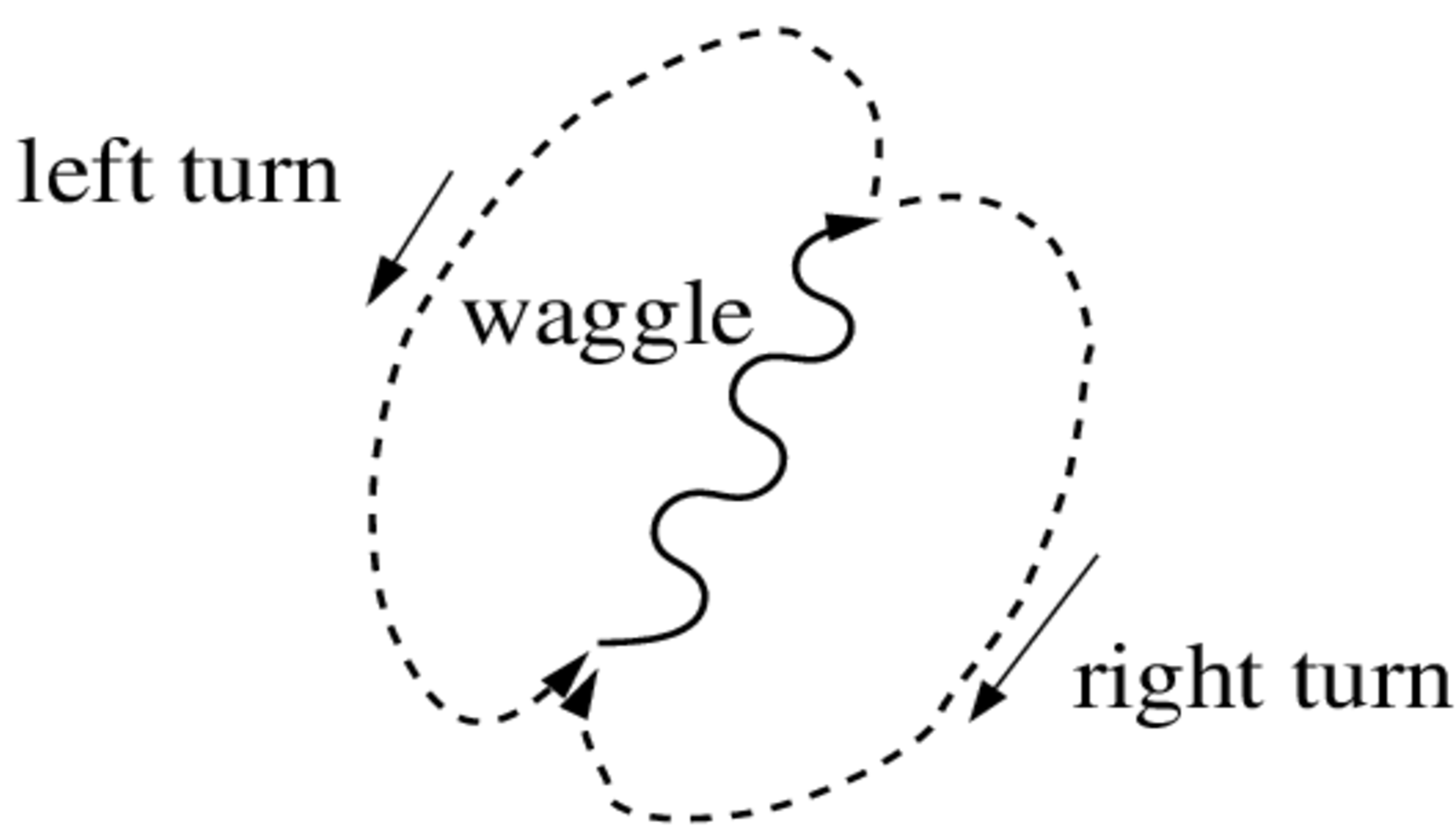}
\label{fig:bee_data_dance_phases}}
\caption{Honey bee dance data. \protect\subref{fig:bee_data_sample} is a sample frame of the original video. \protect\subref{fig:bee_data_dance_phases} is a scheme of dance phases.}
\label{fig:bee_data}
\end{figure}

\section{Numerical experiments with real data}
\label{sec:real_data}

\begin{figure*}[!ht]
\centering
\subfloat{
	\scalebox{1}{\begin{tikzpicture}
\begin{axis}[
  hide axis,
  xmin=0,
  xmax=1,
  ymin=0,
  ymax=1,
  legend columns=3,
  legend style={draw=white!15!black},
  legend cell align={left}
]
\addlegendimage{solid,color={rgb,1:red,0; green,0.4470; blue,0.7410}, line width=1pt}
\addlegendentry{\scriptsize{$x$ coordinates}}
\addlegendimage{dashed,color={rgb,1:red,0.8500; green,0.3250; blue,0.0980}, line width=1pt}
\addlegendentry{\scriptsize{$y$ coordinates}}
\addlegendimage{only marks,mark=x,mark options={fill=black,mark size=4pt, line width = 1pt}, color=black}
\addlegendentry{\scriptsize{true change points}}
\addlegendimage{only marks,mark=x,mark options={mark size=4pt, line width = 1pt}, color={rgb,1:red,0.9290; green,0.6940; blue,0.1250}}
\addlegendentry[text width=100pt,text depth=]{\scriptsize{detections by GP-OLCDT with sliding estimated $\boldsymbol\vartheta_0$}}
\addlegendimage{only marks,mark=x,mark options={mark size=4pt, line width = 1pt}, color={rgb,1:red,0.4940; green,0.1840; blue,0.5560}}
\addlegendentry[text width=100pt,text depth=]{\scriptsize{detections by GP-OLCDT with adaptively estimated $\boldsymbol\vartheta_0$}}
\addlegendimage{only marks,mark=x,mark options={mark size=4pt, line width = 1pt}, color={rgb,1:red,0.4660; green,0.6740; blue,0.1880}}
\addlegendentry{\scriptsize{detections by GP-BOCPD}}
\addlegendimage{only marks,mark=x,mark options={mark size=4pt, line width = 1pt}, color={rgb,1:red,0.3010; green,0.7450; blue,0.9330}}
\addlegendentry{\scriptsize{detections by CUSUM}}
\addlegendimage{only marks,mark=x,mark options={mark size=4pt, line width = 1pt}, color={rgb,1:red,0.6350; green,0.0780; blue,0.1840}}
\addlegendentry{\scriptsize{detections by Contrast}}
\end{axis}
\end{tikzpicture}}
}
\hfil
\subfloat[Sequence 3]{
	\includegraphics[width=0.95\textwidth]{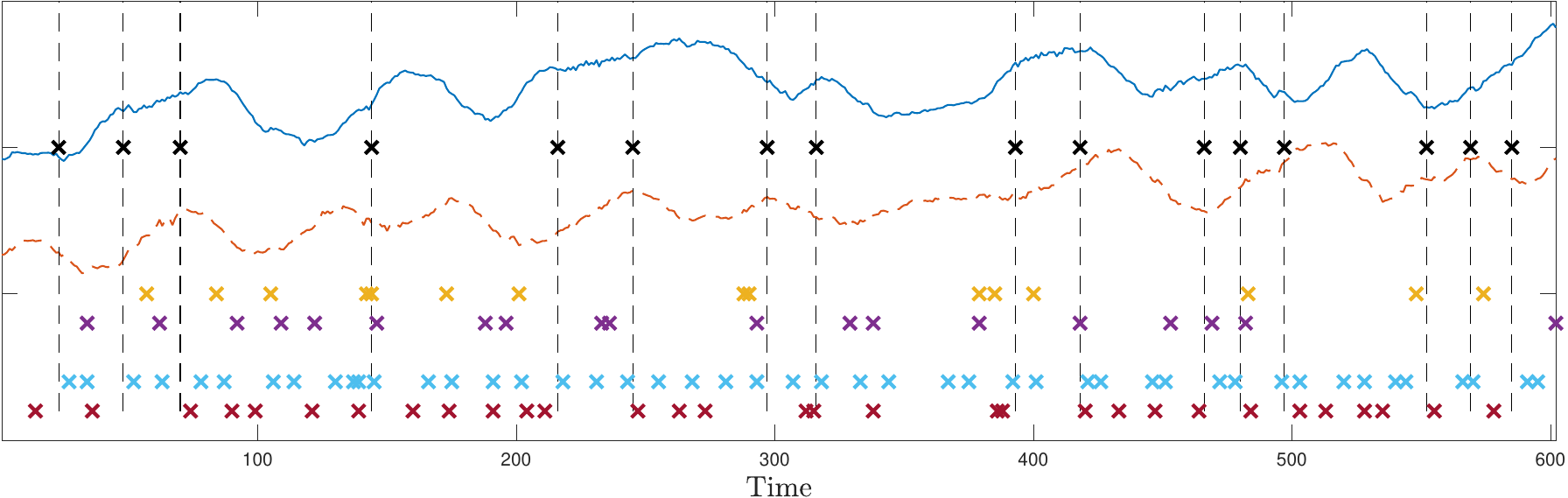}
}
\hfil
\subfloat[Sequence 5]{
	\includegraphics[width=0.95\textwidth]{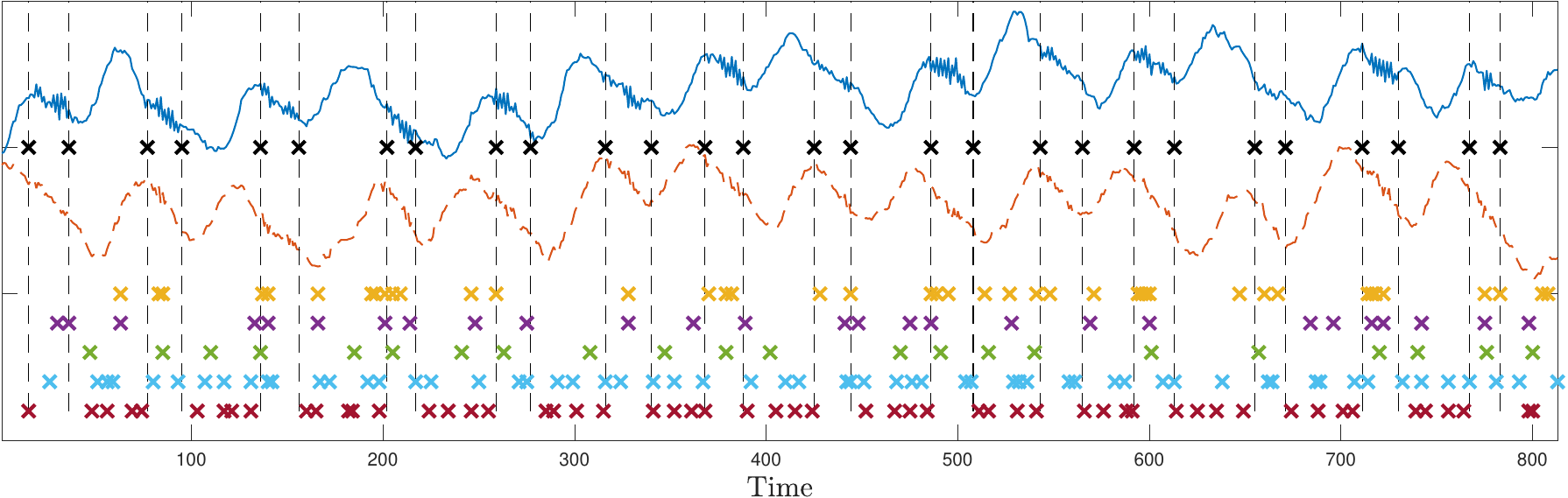}
}
\caption{Change point detection on the honey bee dance data. Two out of total six sequences are presented. $x$ and $y$ coordinates are artificially separated for plotting to increase readability. Vertical lines correspond to true change points. Crosses represent detections by different algorithms.}
\label{fig:bee_data_detections}
\end{figure*}
In this section the proposed framework is applied to the honey bee dance data\footnote{The data is available at \\ \myurl}~\cite{Oh2008}. Honey bees perform a dance as a way to communicate with each other and transfer knowledge about locations of food sources. The dance can be decomposed into three phases: ``left turn'', ``right turn'', and ``waggle'' (Figure~\ref{fig:bee_data}). The original video data provided by the authors~\cite{Oh2008} is pre-processed, where a dancer bee is tracked and its space coordinates and body angle are extracted. The data consists of~$6$ sequences.  

In this paper only spatial coordinates of a tracked bee are processed to find change points between different dance phases. To combine the output from both $x$ and $y$ coordinates of the dancer bee, the methods are applied to both time series independently and then the union of alarms from both time series is used as a resulting output of a method. A function $f$ sampled from a vanilla GP prior is a real-value function, i.e., $f\text{: } \mathbb{R}^n \rightarrow \mathbb{R}$. Although different methods~\cite{Alvarez2011, Nguyen2014, TurnerThesis} have been proposed to overcome this constraint, here scalar-value functions are considered. 

Since the true data model is not known and multiple change points are expected to be detected, the GP-OLCDT with sliding and adaptively estimated hyperparameters are used. The following settings are applied: a constant zero mean function, as input time series are centralised and normalised in advance; a squared exponential covariance function. The sliding window width $\tilde{L}$ for hyperparameter estimation is set to $25$, and the sliding window width $L$ of data for test statistic computation is set to $5$. The significance level $\alpha_{\text{stat}} = 0.01$ is used. The initial values for optimising hyperparameters are set to zero after every change point detection. If there is no change point detected the values of hyperparameters obtained from the previous time step are used as initial values for the current time step.  

The GP-BOCPD method is used with the following settings: a constant zero mean function; a covariance function that is a sum of a rational quadratic, constant and ``white noise'' covariance functions~\cite{TurnerThesis}. The first $250$ data points are used for hyperparameter learning in each sequence. The threshold value for alarm generation is set to $0.99$. 

We also compare with the CUSUM algorithm~\cite{Page1954} and the method based on penalized contrasts~\cite{Lavielle2005, Killick2012}\footnote{The implementation in Matlab 2018a is used (the functions {\tt{cusum}} and {\tt{findchangepts}}, respectively).}(referred below as ``Contrast'') as baselines. To adjust the CUSUM algorithm to multiple change point detection, we restart it after each detected change point, $25$ data points are used to estimate a data mean and standard deviation, a change point is detected when either lower or upper cumulative sums deviates beyond five standard deviations from the estimated mean. The Contrast method is an offline change point detection algorithm. A number of change points and their locations are determined minimising a so-called contrast function. The method is set to detect changes in the mean and the slope and the maximum number of change points is equal to the true number of change points.

The nearest alarm within $5$ time steps before and $15$ time steps after a true change point is treated as a correct detection. All other alarms are processed as false alarms. Samples of the input data and change points detected by the methods are presented in Figure~\ref{fig:bee_data_detections}. The obtained performance measures for all 6 sequences are given in Table~\ref{tab:mean_results_bee_data}.

\begin{table}[!t]
\caption{Change point detection performance on the bee dance data}
\label{tab:mean_results_bee_data}
\centering
\begin{tabular}{p{2.88cm} p{1.33cm} p{0.9cm} p{1.cm} p{0.4cm}}
\toprule
Method     &  Mean time between false alarms  & Mean delay for detection & Missed detection rate & Time (sec)\\
\midrule
\textbf{Sequence 1} & & &\\
GP-OLCDT with sliding estimated~$\boldsymbol\vartheta_0$ & $17.05$ & $\mathbf{2}$ & $0.37$ & $59.94$\\
GP-OLCDT with adaptively estimated~$\boldsymbol\vartheta_0$ & $23.91$ & $2.79$ & $0.26$ & $2.63$\\
GP BOCPD & $\mathbf{25.5}$ & $12$ & $0.84$ & $18.08$\\
CUSUM & $12.31$ & $2.11$ & $\mathbf{0.05}$ & $\mathbf{3}$\textbf{e}$\mathbf{-3}$\\
Contrast& $18.21$ & $3.4$ & $0.21$ & $7$e$-3$\\
\midrule
\textbf{Sequence 2} & & &\\
GP-OLCDT with sliding estimated~$\boldsymbol\vartheta_0$ & $20.11$ & $2.9$ & $0.5$ & $64.13$\\
GP-OLCDT with adaptively estimated~$\boldsymbol\vartheta_0$ & $23.1$ & $3.27$ & $0.5$ & $2.81$\\
GP-BOCPD & $\mathbf{31.75}$ & $5$ & $0.82$ & $18.28$\\
CUSUM & $13.5$ & $2.38$ & $\mathbf{0.05}$ & $\mathbf{3}$\textbf{e}$\mathbf{-3}$\\
Contrast& $17.54$ & $\mathbf{2.24}$ & $0.23$ & $3$e$-2$\\
\midrule
\textbf{Sequence 3} & & &\\
GP-OLCDT with sliding estimated~$\boldsymbol\vartheta_0$ & $\mathbf{28.63}$ & $4.86$ & $0.56$ & $33.78$\\
GP-OLCDT with adaptively estimated~$\boldsymbol\vartheta_0$ & $20.36$ & $5.13$ & $0.5$ & $1.51$\\
GP-BOCPD & --- & --- & $1$ & $17.57$\\
CUSUM & $11.26$ & $2.56$ & $\mathbf{0}$ & $\mathbf{1}$\textbf{e}$\mathbf{-3}$\\
Contrast & $16.5$ & $\mathbf{1.92}$ & $0.19$ & $4$e$-3$\\
\midrule
\textbf{Sequence 4} & & &\\
GP-OLCDT with sliding estimated~$\boldsymbol\vartheta_0$ & $11.5$ & $3.08$ & $0.29$ & $40.21$\\
GP-OLCDT with adaptively estimated~$\boldsymbol\vartheta_0$ & $19$ & $4.2$ & $0.41$ & $1.88$\\
GP-BOCPD & $\mathbf{21}$ & $8.5$ & $0.65$ & $17.82$\\
CUSUM & $13.67$ & $5.18$ & $\mathbf{0}$ & $\mathbf{1}$\textbf{e}$\mathbf{-3}$\\
Contrast & $11.55$ & $\mathbf{3}$ & $0.29$ & $4$e$-3$\\
\midrule
\textbf{Sequence 5} & & &\\
GP-OLCDT with sliding estimated~$\boldsymbol\vartheta_0$ & $9.26$ & $3$ & $0.32$ & $48.07$\\
GP-OLCDT with adaptively estimated~$\boldsymbol\vartheta_0$ & $17.45$ & $3.81$ & $0.43$ & $2.01$\\
GP-BOCPD & $\mathbf{23.67}$ & $6.69$ & $0.43$ & $18.02$\\
CUSUM & $12.57$ & $\mathbf{1.33}$ & $0.14$ & $\mathbf{1}$\textbf{e}$\mathbf{-3}$\\
Contrast & $10.23$ & $1.38$ & $\mathbf{0.07}$ & $9$e$-3$\\
\midrule
\textbf{Sequence 6} & & &\\
GP-OLCDT with sliding estimated~$\boldsymbol\vartheta_0$ & $15.9$ & $4.5$ & $0.6$ & $34.51$\\
GP-OLCDT with adaptively estimated~$\boldsymbol\vartheta_0$ & $\mathbf{17.22}$ & $5.33$ & $0.6$ & $1.45$\\
GP-BOCPD & --- & --- & $1$  & $17.87$\\
CUSUM & $11.97$ & $2.64$ & $\mathbf{0.07}$ & $\mathbf{1}$\textbf{e}$\mathbf{-3}$\\
Contrast & $11.71$ & $\mathbf{-0.08}$ & $0.13$ & $3$e$-3$\\
\bottomrule
\end{tabular}
\end{table}

Overall results show that the CUSUM and Contrast algorithms generate more alarms than GP based methods that leads to low values of all three measures. The GP-BOCPD method rises less alarms and therefore has a good performance in terms of the false alarm measure but also has a very high proportion of missed detections. The proposed GP-OLCDTs demonstrate a tradeoff between the false alarm and missed detection rates. In sequences $3$ and $6$ the GP-BOCPD fails to detect any changes while the GP-OLCDTs are able to detect around $40\%$ of change points. In sequences~$1$ and~$4$ the GP-OLCDTs demonstrate similar results to the offline Contrast method. Analysis of the results given in Figure~\ref{fig:bee_data_detections} confirms that the proposed GP-OLCDTs generate an acceptable number of alarms based on trade-off between the false alarm and missed detection rates.

Table~\ref{tab:mean_results_bee_data} also presents wall-clock time for each of the method given in seconds\footnote{Based on Matlab 2018a implementation on a desktop computer with 4.2 GHz Intel Core i7 CPU, 16 GB RAM}. Both CUSUM and Contrast algorithms are very fast. %and process all sequences within fractions of a second. 
The GP-OLCDT with sliding estimated $\boldsymbol\vartheta_0$ is the slowest method due to a large number of hyperparameter optimisations required. %with each sequence processes by $0.5-1$ minutes. 
The GP-BOCPD is considerably faster than the GP-OLCDT since it optimises hyperparameters only once. % with average computational time for each sequence around $18$ seconds. 
The proposed GP-OLCDT with adaptively estimated $\boldsymbol\vartheta_0$ is a significantly faster method of the proposed online change point detection framework. %, which requires $2-3$ seconds per sequence. 
Since the two versions of the GP-OLCDT have comparable results in terms of performance measures both for the synthetic data and the real data, the faster version with adaptively estimated $\boldsymbol\vartheta_0$ is recommended. Among all compared methods the GP-OLCDT with adaptively estimated $\boldsymbol\vartheta_0$ shows the good trade-off between performance measures and computational time.

%\subsection{Crowd abnormal behaviour video data}

\section{Conclusions}
\label{sec:conclusion}

This paper presents a general framework for change point detection in time series. We considered a data model where a function governing a time series has a Gaussian process prior. A change in time series was defined as a change in hyperparameters of the Gaussian process, which meant a change in a form of functional dependence. 

A change point detection problem was formulated within the statistical hypothesis testing framework. We proposed likelihood ratio based tests for both cases with known and unknown hyperparameters of a Gaussian process before and after a change. We also designed online likelihood-based change point detection tests in a sliding window. 

We analysed the theoretical properties of the developed methods based on the derived probability distribution of the proposed test statistic.

The developed methods were comprehensively evaluated on both synthetic and real data. Under complicated scenarios of the synthetic data the methods demonstrated significantly better empirical performance than the theoretical estimates. On synthetic data that violates the assumptions applied in the tests the proposed methods outperformed the algorithm, which generative model was used in data simulation. The proposed methods were able to detect changes on real data where the existing GP-BOCPD algorithm failed to find any changes. In the sequences where the GP-BOCPD found change points the proposed method demonstrated $52\%$ reduction in the mean delay between the true change time and the time of detection. The proposed method is also $6$ times faster than the GP-BOCPD.  

We investigated the behaviour of the proposed tests under the different observation noise level. It was revealed that for small values of observation noise the tests had the maximum possible power, whereas for the larger noise level the power of the tests varied. Our theoretical analysis of different scenarios for a change showed that the proposed likelihood-based tests have a low power for the more complex cases where the length scale changes from lower to larger values or the signal variance changes from larger to lower values. More generally, the tests would have a low power where data after a change is fitted well into a data model before the change since the tests are based on the likelihood measure. Note that this is the common problem for all change point detection methods based on likelihood such as the contrast and GP-BOCPD methods used as baselines in this paper. Another difficult scenario for a change point detection algorithm is a drift change when, for example, a mean starts to slowly change after a change point. This situation would be also challenging for change point detection methods based on likelihood. Addressing such complicated scenarios provides an interesting avenue for future work.

% % use section* for acknowledgment
% \ifCLASSOPTIONcompsoc
%   % The Computer Society usually uses the plural form
%   \section*{Acknowledgments}
% \else
%   % regular IEEE prefers the singular form
% \fi

% The authors appreciate the support from the EC Seventh Framework Programme [FP7 2013-2017] TRAcking in compleX sensor systems (TRAX) Grant agreement no.: 607400. Lyudmila Mihaylova is also grateful to UK EPSRC for funding this work through EP/T013265/1 project NSF-EPSRC: ShiRAS. Towards Safe and Reliable Autonomy in Sensor Driven Systems. This work was also supported by the USA National Science Foundation under Grant NSF ECCS 1903466.

\ifCLASSOPTIONcaptionsoff
  \newpage
\fi

\bibliographystyle{IEEEtran}
\bibliography{Biblist}

\end{document}